\documentclass[conference]{IEEEtran}
\ifCLASSINFOpdf
\else
\fi

\usepackage[table]{xcolor}
\usepackage{url}
\usepackage{algorithm}
\usepackage{algorithmic}
\usepackage{amsfonts}
\usepackage{bm}
\usepackage{amsmath}
\usepackage{graphicx} 
\usepackage{subfig}  
\usepackage{makecell}
\usepackage{balance}
\usepackage{arydshln}
\usepackage{amssymb}
\usepackage{mathrsfs}
\usepackage{fancybox}
\usepackage{xcolor}
\usepackage{enumitem}
\usepackage{color}
\usepackage{caption}
\usepackage{multirow}

\begin{document}
%
\title{Enhanced Attacks on Defensively Distilled \\Deep Neural Networks}

\author{\IEEEauthorblockN{Yujia Liu, Weiming Zhang, Shaohua Li, Nenghai Yu}
\IEEEauthorblockA{University of Science and Technology of China\\
Email: yjcaihon@mail.ustc.edu.cn}}

\maketitle


\begin{abstract}
Deep neural networks (DNNs) have achieved tremendous success in many tasks of machine learning, such as the image classification. Unfortunately, researchers have shown that DNNs are easily attacked by \textit{adversarial examples}, slightly perturbed images which can mislead DNNs to give incorrect classification results. Such attack has seriously hampered the deployment of DNN systems in areas where security or safety requirements are strict, such as autonomous cars, face recognition, malware detection. Defensive distillation is a mechanism aimed at training a robust DNN which significantly reduces the effectiveness of adversarial examples generation. However, the state-of-the-art attack can be successful on distilled networks with 100\% probability. But it is a white-box attack which needs to know the inner information of DNN. Whereas, the black-box scenario is more general. In this paper, we first propose the \textit{$\epsilon$-neighborhood attack}, which can fool the defensively distilled networks with 100\% success rate in the white-box setting, and it is fast to generate adversarial examples with good visual quality. On the basis of this attack, we further propose the \textit{region-based attack} against defensively distilled DNNs in the black-box setting. And we also perform the \textit{bypass attack} to indirectly break the distillation defense as a complementary method. The experimental results show that our black-box attacks have a considerable success rate on defensively distilled networks.
\end{abstract}


%
\IEEEpeerreviewmaketitle

\section{Introduction}
Deep Neural Networks (DNNs) have led to major breakthroughs in recent years and have been powerful tools in various applications, including computer vision \cite{krizhevsky2012imagenet,szegedy2015going,simonyan2014very}, speech recognition \cite{hinton2012deep}, natural language processing \cite{mikolov2010recurrent}, healthcare \cite{kriegeskorte2015deep,liang2014deep}, \textit{etc.} Despite DNNs' huge success, researchers have discovered that they are vulnerable to deliberate attacks. An attacker can generate a slightly modified sample, called \textit{adversarial example} \cite{szegedy2013intriguing}, to mislead a DNN to give an incorrect output. This kind of attack is particularly horrible in security-critical systems, such as autonomous cars \cite{bojarski2016end}, face recognition \cite{parkhi2015deep}, malware detection \cite{dahl2013large}. But the existence of attacks can motivate more researches about how to defense adversarial examples and get a more robust DNN.

Many studies focus on attack methods, that is, how to generate adversarial examples. In image classification task, some of these attacks are based on the gradient of network, such as FGSM (Fast Gradient Sign Method) \cite{goodfellow2014explaining}, FGV (Fast Gradient Value) \cite{rozsa2016adversarial}, JSMA (Jacobian Saliency Map Attack) \cite{papernot2016limitations}, and some are based on solving optimization problems using different methods such as L-BFGS \cite{szegedy2013intriguing}, Deepfool \cite{moosavi2016deepfool}, C\&W attack (by Carlini \& Wagner) \cite{carlini2017towards}. They have their own merits and defects in attack success rate, runtime, visual quality of generated adversarial examples.
\begin{figure}[!ht]
	\centerline{\textbf{CIFAR10}}
	\centerline{original\hspace{2.5cm}adversarial examples\hspace{1.4cm}}
	\vfill
	\vspace{3pt}
	\begin{minipage}[t]{0.13\textwidth}
		\includegraphics[height=2cm,width=2cm]{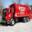}
	\end{minipage} 
	\hfill  
	\begin{minipage}[t]{0.11\textwidth}
		\includegraphics[height=2cm,width=2cm]{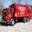}  
	\end{minipage} 
	\hfill  
	\begin{minipage}[t]{0.11\textwidth}
		\includegraphics[height=2cm,width=2cm]{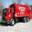}  
	\end{minipage} 
	\hfill  
	\begin{minipage}[t]{0.11\textwidth}
		\includegraphics[height=2cm,width=2cm]{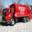}  
	\end{minipage} 
	\vfill
	\vspace{4pt}
	
	\begin{minipage}[t]{0.13\textwidth}
		\includegraphics[height=2cm,width=2cm]{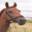}
	\end{minipage} 
	\hfill  
	\begin{minipage}[t]{0.11\textwidth}
		\includegraphics[height=2cm,width=2cm]{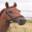}  
	\end{minipage} 
	\hfill  
	\begin{minipage}[t]{0.11\textwidth}
		\includegraphics[height=2cm,width=2cm]{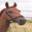}  
	\end{minipage} 
	\hfill  
	\begin{minipage}[t]{0.11\textwidth}
		\includegraphics[height=2cm,width=2cm]{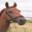}  
	\end{minipage} 
	\vfill
	\vspace{5pt}
	\centerline{\textbf{ImageNet}}
	\centerline{\hspace{0.5cm}original\hspace{2.7cm}adversarial example}
	\vfill
	\vspace{3pt}
	\begin{minipage}[t]{0.24\textwidth}
		\includegraphics[height=4.3cm,width=4.3cm]{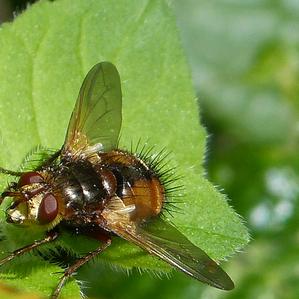}  
	\end{minipage} 
	\hfill  
	\begin{minipage}[t]{0.24\textwidth}
		\includegraphics[height=4.3cm,width=4.3cm]{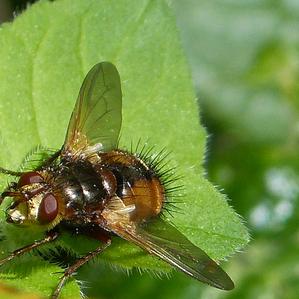}  
	\end{minipage} 
	\caption{An illustration of our generated adversarial examples for RGB images in CIFAR10 on distilled network and in ImageNet on Inception-v3. In the first two row, the original labels of the two images are \{truck, horse\}, and the three columns of adversarial examples correspond to three randomly selected randomly selected target classes \{airplane, automobile,bird\}. The original class of the  ImageNet image in the third row is \{fly\}, and the target class is \{house finch\}. Results show that the adversarial perturbations are imperceptible by human eyes.} 
	\label{fig:intro-show}		
\end{figure}

Many researchers have proposed various defense methods against above attacks. These defenses try to make DNNs still give the correct output on adversarial examples or make the success rate of generating adversarial examples decrease significantly. The basic ideas are roughly divided into two categories. One is based on image preprocessing \cite{lu2017no, das2017keeping, cao2017mitigating}, taking advantage of the spatial instability of adversarial examples. Before feeding an input image into a DNN, people will perform some operation on it in the spatial domain, such as adding noise, JPEG compression, cropping, rotation, \textit{etc}. And other ideas focus on getting a new robust DNN for adversarial examples by modifying the network architecture or the training process \cite{gu2014towards, papernot2016distillation, kurakin2016adversarial}. 

Defensive distillation \cite{papernot2016distillation} is a strong defensive mechanism based on training a new robust network by using the distillation method. It can significantly reduce the effectiveness of adversarial examples on DNNs. But Carlini and Wagner propose a powerful attack (C\&W attack) that can still acquire 100\% success rate on DNNs trained with defensive distillation \cite{carlini2017towards}, which is the state-of-the-art attack. However, C\&W attack still has room for improvement. First, it needs a long runtime to generate an adversarial example. And besides, it is a white-box attack, that is, attackers need to know the output inside the network. In the actual adversarial environment, the black-box attack is more practical, that is, attackers only know the output probability vector of the classification result.

In this paper, we first propose an $\epsilon$-neighborhood attack on defensively distilled networks in the white-box setting, which, as we will illustrate below, achieves high visual quality and is able to generate adversarial examples of any target class very fast. On this basis, we further propose a novel region-based attack on distilled networks in the black-box setting. To the best of our knowledge, this attack is the first attempt to generate adversarial examples with only access to the output classification probabilities of a distilled network.

Given an image, our $\epsilon$-neighborhood attack limits the upper bound of perturbation $\epsilon$ to each pixel, thus facilitating the control of modification to original images, and ensuring no obvious bright spots appear in the generated adversarial examples. On ImageNet \cite{deng2009imagenet} classification task, our attack can cause the Inception-v3 \cite{szegedy2016rethinking} to misclassify all the image into target class, and the largest perturbation to each pixel is even no more than 2. Actually, our large-scale experiment shows that we can achieve over 70\% success rate even if $\epsilon$ is fixed to 1, for example, as shown in Fig.\ref{fig:intro-show}. Human eyes cannot completely tell the difference between the adversarial examples and the original ones. In addition, our method is fast to execute. And if the preset upper bound $\epsilon$ goes up, it can be even faster.

As we mentioned above, the state-of-the-art attack (C\&W attack) can only against defensively distilled networks in the white-box setting. However, in many real-world cases, such as machine-learning-as-a-service (MLaaS), an attacker only has access to the final output probabilities for an image, that is to say, he cannot get any results of the inner layers. In this black-box setting, distilled networks can against all the published attacks. It seems that the defensive mechanism is very effective. However, our proposed region-based attack can succeed in finding adversarial examples on distilled networks. And we then introduce the bypass attack to guarantee a high success rate under different conditions (various distillation temperatures of the distilled models), which indirectly breaks the distillation defense. We will show that once a low-temperature model is accessible, its corresponding enhanced models with high temperatures will be no longer unsecure and easily attacked by our bypass method.

We make the following contributions in this paper:
\begin{itemize}
	\item We propose an $\epsilon$-neighborhood attack to defensively distilled networks with $100\%$ success rate in the white-box setting. This attack limits the maximum perturbation, denoted as $\epsilon$, of each pixel in an image, thus ensuring a good visual quality of the generated adversarial examples. Our experiments show that this attack is fast to execute.
	\item We present a region-based attack on defensively distilled networks in the black-box setting, which, to our best knowledge, is the first  attemp. Better still, our generated adversarial examples are robust to noise.
	\item We further introduce bypass attack on the basis of our region-based attack as a complementary method in the black-box setting. We provide experimental results to indicate that by the bypass attack, any enhanced models with high distillation temperature for security concern will become unsecure with the assistance of a low-temperature model.
\end{itemize}

\section{Preliminaries}
\label{Preliminaries}

\subsection{Notations for Deep Neural Networks}
\label{Notations-for-Deep-Neural-Networks}
In this paper, we focus on DNNs for image m-classification task. A trained DNN is represented as $f:\mathbf{x}\rightarrow \mathbf{y}$, where $\mathbf{x}\in \mathbb{R}^{n}$ is an input normalized image and $\mathbf{y}\in \mathbb{R}^{m}$ is an output vector, representing the probabilities of the image belonging to each class. The raw input is a grayscale image or a three-dimensional RGB image, and here we scale every pixel from range [0,255] to [0,1]. The output vector $\mathbf{y}=(y_{1},y_{2},...,y_{m})$ satisfies $\sum_{i=1}^m y_i=1$, and $y_{i} \in [0,1]$, $i=1,2,...m$. We define $f$ as the full neural network, including several hidden layers for extracting features and a softmax layer for mapping real output values of a network to a probability vector,
\[
f(\mathbf{x})=\text{softmax}(f_{n}(f_{n-1}(...f_1(x)...))) = \mathbf{y},
\]
where $f_{i}$ is the inner layer function.
$Z(\mathbf{x})=\mathbf{z}=(z_{1},z_{2},...z_{m}) \in \mathbb{R}^{m}$ is the pre-softmax result vector (called \textit{logits}), that is, $\mathbf{y}=$softmax$(Z(\mathbf{x}))$. The softmax function is as follows:
\[\text{softmax}(\mathbf{z})_{i}=\frac{e^{z_{i}}}{\sum_{i=1}^m e^{z_{i}}}, \quad 1 \leq i \leq m.
\]

\subsection{Targeted Attack and Untargeted Attack}
Targeted attack is aimed at producing adversarial examples that force the DNN to misclassify them into a specific target class. For an original image $\mathbf{x}$ with its original true label $C(\mathbf{x})=\mathop{\arg\max}_{i}y_{i}$, the attacker chooses a particular target class $C_{t}\neq C(\mathbf{x})$, and a targeted attack is to find an adversarial example $\mathbf{x'}$, which is close to $\mathbf{x}$, satisfying its result label $\mathop{\arg\max}_{i}y_{i}'=C_{t}$.

The untargeted attack is to make the DNN classify the produced adversarial example into any other class except the original one. Attackers try to find $\mathbf{x'}$ so that $\mathop{\arg\max}_{i}y_{i}'\neq C({\mathbf{x}})$. Previous researches have shown that an untargeted attack is easier to succeed. So in this paper, we focus on the more challenging targeted attack.

\subsection{White-box Attack and Black-box Attack}
According to how much information attackers know about the DNN, attacks can be divided into white-box and black-box attack. In the white-box setting, attackers have all the knowledge about the architecture, parameters, training set of the network, and intermediate results of inner layers (\textit{e.g.} logits). It is a more favorable situation for attackers to produce adversarial examples. 

As for the black-box, attackers do not have knowledge of the network. They only have access to the output probability vector $\mathbf{y}$ of the DNN. They can query the DNN for any input according to their needs. It is more difficult for attackers to generate adversarial examples. In this paper, we discuss attacks of both two settings above.

\subsection{Success Rate of Attack}
In a targeted attack, a successful adversarial example can mislead the network to classify it into the target class. Here we define the success rate as the proportion of adversarial examples, whose classification result is the target label, to all our test images, regardless of whether the original images classification results are correct or not.

\section{Known Attacks and Defenses}
\label{Known-Attacks-and-Defenses}
\subsection{Known Attacks}
\subsubsection{FGSM}
Fast gradient sign method \cite{goodfellow2014explaining} is a gradient-based fast algorithm of computing adversarial perturbation with a $\ell_{\infty}$ constraint. It explores the gradient direction of the loss function $\text{Loss}(\mathbf{x},t)$ and adds a fixed perturbation to the original image. Attackers can get  adversarial example $\mathbf{x'}$ from:
\[\mathbf{x'}=\mathbf{x}-\tau \cdot \text{sign}(\nabla_{\mathbf{x}}\text{Loss}(\mathbf{x},t)),
\] where $\tau$ is usually a small value to control the modification magnitude. $\text{Loss}(\mathbf{x},t)$ represents the cost of classifying $\mathbf{x}$ into the target class $t$. FGSM is just designed to be computationally efficient rather than an optimal attack.
\subsubsection{JSMA} 
Jacobian saliency map attack \cite{papernot2016limitations} is another gradient-based method, which uses the calculated saliency map to modify one pixel with the highest saliency maps value in each iteration, until the classification result turns to the target class. This method modifies images more precisely at the expense of a greater computing cost than FGSM.

\subsubsection{Deepfool} Deepfool \cite{moosavi2016deepfool} generates an adversarial example by exploring the nearest decision boundary and crossing it to deceive the network. It is an untargeted attack. It is usually difficult to solve the problem directly, so it assumes DNN is linear with hyperplanar decision boundary. The optimal update direction is obtained step by step in an  iterative way. The image is modified a little to reach the boundary in each iteration, until reaching and going across it. 

\subsubsection{C\&W}
\label{CW}
C\&W attack \cite{carlini2017towards} is the state-of-the-art white-box attack at the time of our work. The attack can be targeted or untargeted and has three forms based on different distortion measures ($\ell_{2}$, $\ell_{0}$, $\ell_{\infty}$). Its $\ell_{2}$ form has the best performance. In this paper, all the referred C\&W attack in the following sections is its $\ell_{2}$ norm. It constructs adversarial examples by solving the following optimization problem:
\begin{equation}
\label{CW-problem}
\begin{aligned}
&\mathop {\mathrm{minimize}}\limits_\delta \left \| \delta \right \|_{2}+c\cdot l(x+\delta)\\
&\quad\text{s.t.}\quad x+\delta \in [0,1]^{n}
\end{aligned}
\end{equation}
The objective is to find the smallest perturbation measured by $\ell_{2}$ norm (represented by the first term $\left \| \delta \right \|_{2}$) and it can make the network misclassify it into the target class (represented by the second term $l(x+\delta)$). The loss function $l(\cdot)$ reflects the distance between the current situation and the objective of attack and is defined as:
\begin{equation*}
l(x)=\text{max}(\text{max}_{i\neq t}\{Z(x)_{i}\}-Z(x)_t,-\kappa)
\end{equation*}
$Z(x)$ is the logits mentioned in Section \ref{Preliminaries}, $t$ is the target label, and $\kappa$, called confidence, is a hyper-parameter for enhancing the transferability to other models, which is discussed detailedly in \cite{carlini2017towards}. Under normal circumstances, attackers can set $\kappa=0$. When $\text{max}_{i\neq t}\{Z(x)_{i}\}-Z(x)_t>0$, it means the adversarial example $x$ is not classified into the target class, so it is not a successful attack. When $\text{max}_{i\neq t}\{Z(x)_{i}\}-Z(x)_t\leq 0$, it implies a successful adversarial example. In addition, C\&W attack turns an optimization problem with a box constraint $x+\delta \in [0,1]^{n}$, into an unconstrained problem by replacing $\delta$ with $\frac{1}{2}(\tanh (w)+1)-x$, where $w$ is a new optimizer ranging in $(-\infty, +\infty)$. So various optimization algorithms which are only suitable for unconstrained problems can be used.

\subsubsection{ZOO} Zeroth Order Optimization attack \cite{chen2017zoo} is a black-box attack inspired by C\&W. It substitutes $\log f(x)$ for logits $Z(x)$ in the loss function, so attackers just need to query the final classification vector $f(x)$. And ZOO can solve the new optimization problem by stochastic coordinate descent method, ADAM optimizer, or Newton’s method. In addition, to make the attack more efficient, it reduces the attack-space dimension by resizing the perturbation from high to low. However, this attack is not designed for defensively distilled networks.

\subsection{Known Defenses}
\subsubsection{Detecting adversarial examples}
Many reseaches focus on detecting adversarial examples \cite{lu2017safetynet, metzen2017detecting, carlini2017adversarial,meng2017magnet}. A detector needs to distinguish whether the input image is an adversarial example or not, so it is a binary classification problem. Once the image is detected to be adversarial, it may be manually labeled. But if so, the entire system won't be fully automatic. Hence, only detecting adversarial examples but not trying to classifying it into the correct class, is a weak defense mechanism.

\subsubsection{Pre-processing input}
It is a simple method that pre-processing all the input images before feeding them into a classifier to eliminate adversarial perturbations. This utilizes that adversarial examples are not spatially robust. When defenders interfere with them, they may recover to the original class and not be adversarial examples anymore. Some common operations are adding noise, JPEG compression \cite{das2017keeping}, cropping, rotating, \textit{etc.} In addition, some researchers use principal component analysis to reduce dimension for input images and remove adversarial perturbations \cite{bhagoji2017dimensionality}. And autoencoders are also used for reconstructing the adversarial example in order to make it recover to the original label \cite{meng2017magnet}.

\subsubsection{Adversarial training}
Adversarial training \cite{kurakin2016adversarial} is aimed at training a new robust network by adding adversarial examples into the training set. Defensers train a network using both normal images and adversarial examples, so that the network learns the features from both of them and can classify adversarial examples correctly. However, adversarial training method is not robust to strong attacks like C\&W and our work, it cannot recognize adversarial examples that do not appear during training \cite{cao2017mitigating}.
\subsubsection{Defensive distillation}
\label{defensive-distillation}
Defensive distillation \cite{papernot2016distillation} is a strong defense mechanism of training a robust network which makes it almost impossible for many gradient-based attacks to generate adversarial examples on it, because its output probabilities is too ``hard''. In this defense, the softmax function is modified as follows,
\[F(\mathbf{x})=\left[\frac{e^{z_{i}/T}}{\sum_{i=1}^m e^{z_{i}/T}}\right]_{i\in0...m},
\]
where $T$ is the distillation temperature. First, it trains a teacher model with the new softmax function, and then for each sample in the training set, feeds them into the teacher model to obtain its corresponding ``soft'' label. After that, it trains the distilled network on the soft labels.
%
\begin{figure}[h]
``soft'': \hspace{3pt}\fbox{0.0001, 0.0009, 0.0031, 0.9704, 0.0004,..., 0.0000}
\vfill
\vspace{2pt}
``hard'': \fbox{0.0000, 0.0000, 0.0000, 1.0000, 0.0000,..., 0.0000}
\end{figure}

Soft and hard labels are just shown as above for example. The trained distilled network will then have a good defensive performance. 

\section{Our $\epsilon$-neighborhood Attack}
\label{Our-e-neighborhood-Attack}
In this section, we present our algorithm in details. Our proposed $\epsilon$-neighborhood attack is aimed at finding an adversarial example, in which the perturbation to each pixel of original image does not exceed $\epsilon$. The hyper-parameter $\epsilon$ can be considered as the perturbation intensity that attackers can tolerate. It is an important adjustable parameter controlled by attackers as needed to balance the running speed and visual quality. Here, we will describe our new objective function within the constraint of $\epsilon$, and explain why it has good performance in distillation resistance, speed, and visual quality.

\subsection{Objective Function}
Our goal is to search an adversarial perturbation $\delta$ with an upper limit $\epsilon$ for a normalized image $x$ to turn it into a target label $t$, and the perturbed image $x'$ still needs to be a valid image. That is to say, we try to find an adversarial example in the $\epsilon$-neighborhood of the original image. It is described as follows:
\begin{equation}\label{ori_problem}
\begin{aligned}
\text{find} \ \delta, \ \ \text{s.t.} \ 
&\mathop{\arg\max}f(x+\delta)=t\\
&\ \delta \in [-\epsilon,\epsilon]^{n}\\
&\ \delta+x \in [0,1]^{n},
\end{aligned}
\end{equation}
where $x\in[0,1]$ is fixed, and $\epsilon\in[0,1]$ is a fixed but attacker-adjustable hyper-parameter. It is hard to solve the problem above, so we usually convert it to an optimization problem by introducing loss functions. So we move on to the problem:
\begin{equation}
\label{Our-optimization-problem}
\begin{aligned}
&\mathop {\mathrm{minimize}}\limits_\delta \mathit{L}_{t}(x+\delta),\\ 
&\ \text{s.t.} \quad \delta \in [-\epsilon,\epsilon]^{n}\\
&\qquad\ \;\!\delta+x \in [0,1]^{n}.
\end{aligned}
\end{equation}
Here, the loss funtion $\mathit{L}_{t}(\cdot)$ we used in our $\epsilon$-neighborhood attack is inspired by C\&W \cite{carlini2017towards}, which is defined as
\begin{equation}
\label{CW-loss-function}
\mathit{L}_{t}(x)=\text{max}(\text{max}_{i\neq t}\{Z(x)_{i}\}-Z(x)_t,-\kappa).
\end{equation}
The details have been mentioned in Section \ref{CW}.

Our goal is to reduce $\mathit{L}_{t}(x+\delta)$ to below 0. If so, the logit of the target class is the largest, and so is its corresponding probability. Therefore the output is our target label. And the attack will be successful.

Compared with the optimization problem (\ref{CW-problem}) mentioned in Section \ref{CW}, our problem (\ref{Our-optimization-problem}) is different. We do not minimize the term $\left \| \delta \right \|_{2}$, instead, we only minimize the loss function directly. We use an adjustable parameter $\epsilon$ to control how much perturbation will be added. The experiments in Section \ref{Performance-of-neighborhood-Attack} will show that this parameter can guarantee image visual quality very well. Better still, our optimization objective with only one term (previous work all two terms) improves the processing speed greatly.

Our optimization problem is box-constraint, for which many optimizers are not suitable. But previous work has presented several solutions, like L- BFGS-B, projected gradient descent, variable substitution, \textit{etc.} Here we take the idea of variable substitution, that is, we turn the box-constraint problem to an unconstrained optimization. Its difficulty lies in our optimization has two constraints, we transform the original constraints to the following inequality:
\begin{equation}\label{max-min}
\max(-\epsilon+x,0)\leq \delta+x \leq \min(\epsilon+x,1)
\end{equation}
We use  $\tanh(\cdot)$ to substitute for the original variable $\delta$ with a new variable $w$. To respect the constraint of inequality (\ref{max-min}), we derive the following equation: 
\begin{equation*}
\begin{aligned}
&\delta=\frac{b-a}{2}\tanh{w}+\frac{b+a}{2}-x,\\
\text{where,}\quad &a=\max(-\epsilon+x,0), \ b=\min(\epsilon+x,1).\\
\end{aligned}
\end{equation*}
Since $-1\leq\tanh{w}\leq1$, we have
$$a\leq\delta+x\leq b,$$
which is the same as inequality (\ref{max-min}). Hence, the new optimization problem we focus on is as follows:
\begin{equation}
\label{final-formula}
\mathop {\mathrm{minimize}}\limits_w \mathit{L}_{t}(\frac{b-a}{2}\tanh{w}+\frac{b+a}{2}).
\end{equation}
Formula (\ref{final-formula}) is our final objective function.

Compared with the original problem (\ref{Our-optimization-problem}), there is no constraint in our new optimization problem. Note that, given $\mathbf{x}$ and $\epsilon$, $a$ and $b$ will be determined immediately. Thus we have an unconstraint optimization problem\footnote{In our actual execution, we add another term $c$ on the $w$ to balance the optimization, where $c=-\text{arctanh}(\frac{b+a-1}{b-a})$. So the optimization function is $\mathop {\mathrm{minimize}}\limits_w \mathit{L}_{t}(\frac{b-a}{2}\tanh{(w+c)}+\frac{b+a}{2}).$ And our experiment shows that it will make a little contribution to the processing speed.}.
There have many optimizers to choose, such as Gradient Descent Optimizer, Adagrad Optimizer, Momentum Optimizer, Adam Optimizer, \textit{etc.} After our experiments on them, we adopt Adam as the optimizer to solve the new problem (\ref{final-formula}), which can find desired $w$ with the fastest convergence speed.

\subsection{Distillation Resistance}
As described in Section \ref{defensive-distillation}, using distillation in the training process will make the network more robust. When feeding an image $x$ into a distilled network, the output classification vector $f(x)$ is ``hard", which means its largest probability is 1.0 or very close to 1.0, and others are all 0 or nearly 0. This makes the gradient of $f(x)$ is also almost always 0. Therefore the attacks based on gradient are deactivated, because they cannot get effective gradient value to compute the needed adversarial perturbation (like FGSM attack) or solve the optimization problem (like L-BFGS, JSMA attack). However, our $\epsilon$-neighborhood attack and previous C\&W attack can resist distillation due to the fact that we use the logits before the softmax layer instead of the probability, so the ``hard" output has no effect on us. What we should not ignore is that there still exist some method which uses logits but cannot resist distillation. This is because the distilled network also causes considerable variation of the output logits, called ``hard" logits (the large values become much larger and the small values become much smaller), it is so difficult to find a kind of perturbation to change the rankings of logits directly. However, the loss function of our $\epsilon$-neighborhood attack is the difference between the maximum value of other logits (except target class logits) and the target class logits. So what we care are only two logits values. No matter how big the disparity there becomes between these logits, it is possible to make the difference between the two values very small to 0. This has nothing to do with ``hard" logits caused by distillation. Therefore, our $\epsilon$-neighborhood attack is a successful one to resist distilled network.

\subsection{Speed Improvements}
Besides distillation resistance, our attack greatly increases the speed of generating adversarial examples compared with C\&W. Although we are using the same optimizer (Adam), our runtime of producing one adversarial example is several or even tens of times shorter than that of C\&W. Almost all previous researches focus on minimize the perturbation (maybe based on different norms: $\ell_{0}$, $\ell_{2}$, $\ell_{\infty}$) as one term of the objective function in optimization. And another term is always the loss function measuring the success of attack. Just as in C\&W, its objective is $\mathop {\mathrm{minimize}}\limits_\delta \left \| \delta \right \|_{2}+c\cdot l(x+\delta)$. It is the sum of two terms, and $c$ is a balancing weight. In the optimization process, C\&W has to find an optimal $c$ by binary search, which consumes a long time. However, in our $\epsilon$-neighborhood attack, we do not need to minimize the perturbation as one optimization term, and we integrate the two constraints (maximum perturbation constraint and image validity constraint) into the final loss function. This makes our function has only one term, so it is easier to solve the problem directly by optimizer without the need of auxiliary calculation. 

\subsection{Visual Quality}
Although our objective function does not minimize the total perturbation, it still ensures good visual quality of the adversarial examples. Intuitively, several large perturbations to pixels are spread to the entire image by  presetting the upper bound. The experiment will show that we can always get a lower maximum adversarial perturbation on an image than C\&W attack on the MNIST \cite{lecun1998mnist}, CIFAR10 \cite{krizhevsky2009learning} and ImageNet \cite{deng2009imagenet} dataset. Its immediate benefit is that the adversarial examples generated by our $\epsilon$-neighborhood attack will absolutely not have 
obvious bright spots, which can cause the most damage on the  visual effect of images. This is more in line with the requested perceptually invisibility of adversarial examples. And better still, on the ImageNet dataset, our adversarial perturbations sometimes have even smaller total distortion than C\&W. Recall that C\&W attack needs to binary search the optimal balance weight $c$, but it cannot find the proper $c$ all the time.

\section{Attacks on Black-box Distilled Model}
\label{Attack-on-Black-box-Distilled-Model}
In this section, we describe our attacks on the black-box distilled model. One limitation of C\&W attack is that it is a white-box attack. It has to use the output of the inner layer (logits) during its optimization. As for the black-box attack, however, what we can use is only the output probabilities. To the best of our knowledge, it is the first attempt to implement black-box attack against defensively distilled networks.

\subsection{Substitute Probability for Logits}
In order to avoid the use of logits, we need to find a substitute in the loss function (\ref{CW-loss-function}). Our idea is replacing logits $Z(x)$ with $\log f(x)$. Then the loss function is:
\begin{equation}
\label{log-loss-function}
\tilde{\mathit{L}_{t}}(x)=\text{max}(\text{max}_{i\neq t}\{\log f(x)_{i}\}-\log f(x)_t,-\kappa)
\end{equation}
We can derive it is exactly equivalent to the loss function (\ref{CW-loss-function}) with logits. As mentioned in Section \ref{Notations-for-Deep-Neural-Networks}, 
\[f(x)_{i}=\frac{e^{Z(x)_{i}}}{\sum_{i=1}^m e^{Z(x)_{i}}}, \quad 1 \leq i \leq m
\]
So the difference between two log value is 
\begin{equation*}
\begin{aligned}
&\log f(x)_{i}-\log f(x)_{j}\\
=&(Z(x)_{i}-\log \sum_{i=1}^m e^{Z(x)_{i}})-(Z(x)_{j}-\log \sum_{j=1}^m e^{Z(x)_{j}})\\
=&Z(x)_{i}-Z(x)_{j}, \qquad\qquad 1 \leq i,j \leq m
\end{aligned}
\end{equation*}
Therefore, we can conclude that the loss function $\tilde{\mathit{L}_{t}}(x)$ only with $f(x)$,  is equal to  $\mathit{L}_{t}(x)$ using logits $Z(x)$. We do not need to know the  internal information of network anymore during the optimization process, but just query the final output probabilities. By substituting the $\log$ value of probability for logits, black-box attack can be achieved.

\subsection{Our Region-based Attack}
\label{Our-Region-based-Attack}
Defensively distilled networks lead to such a ``hard" output vector for any input image that generating adversarial examples becomes very difficult. And for a black-box attack, what we can only use is the ``hard" vector. The ``hard'' means the largest output probability is 1.0 or very close to 1.0, and others are all 0 or nearly 0.  We find it still difficult for us to solve the optimization problem on distilled networks, even if we use the black-box loss function (\ref{log-loss-function}). Our experiment shows that in the optimization procedure, ``hard" output leads to the initial loss value is $\infty$, and after several iterations, the loss converges to a positive number, which indicates that the attack is not successful. Intuitively, if we assume an image as a point in the high-dimensional feature hyperspace, then distillation will make the point very ``stubborn''. No matter how we try to move the point towards the direction of the target class, it still cannot get out of its original class region.

To solve the problem mentioned above, we transform the optimization on a data point to lots of points in a region in the hyperspace. More specifically, in each optimization iteration, we add Gaussian noise on the current image to make its corresponding point in hyperspace slightly deviate from its position. Then we calculate the loss of the noise-added image, and minimize it by the Adam optimizer. After many iterations, it will eventually find a route, along which the data point will rush out of its original class region into the target region in the hyperspace. The process is described in Algorithm \ref{Region-based-Attack-algorithm}.

\begin{algorithm}[h]
	\caption{Region-based Attack on Black-box Classifier}
	\label{Region-based-Attack-algorithm}
	\begin{algorithmic}
		\STATE \textbf{Input:} Original image $x$, iteration number $N$, maximum perturbation $\epsilon$.
		\STATE \textbf{Output:} Adversarial example $x'$.
		\STATE \textbf{Initialize:} $x_{0}$ $\Leftarrow$ $x$, $i$ $\Leftarrow$ $0$
		\WHILE {$i<N$ \&\& $\tilde{\mathit{L}_{t}}(x;\epsilon)>0$}
		\STATE $x_{i}^{\text{noise}}\Leftarrow x_{i}+ \mathcal{N}(0,\sigma^{2})$;
		\STATE Calculate $\tilde{\mathit{L}_{t}}(x;\epsilon)$;
		\STATE Get $\delta_{i}$ according to $\tilde{\mathit{L}_{t}}(x;\epsilon)$ by Adam optimizer;
		\STATE Update $x_{i+1}\Leftarrow x_{i}+\delta_{i}$;
		\STATE \quad\quad\quad$\,$ $i\Leftarrow i+1$
		\ENDWHILE
		\STATE $x'\Leftarrow x_{i}$
		\RETURN $x'$
	\end{algorithmic}
\end{algorithm}

\begin{figure}
	\centering
	\includegraphics[height=2.3in, width=3in]{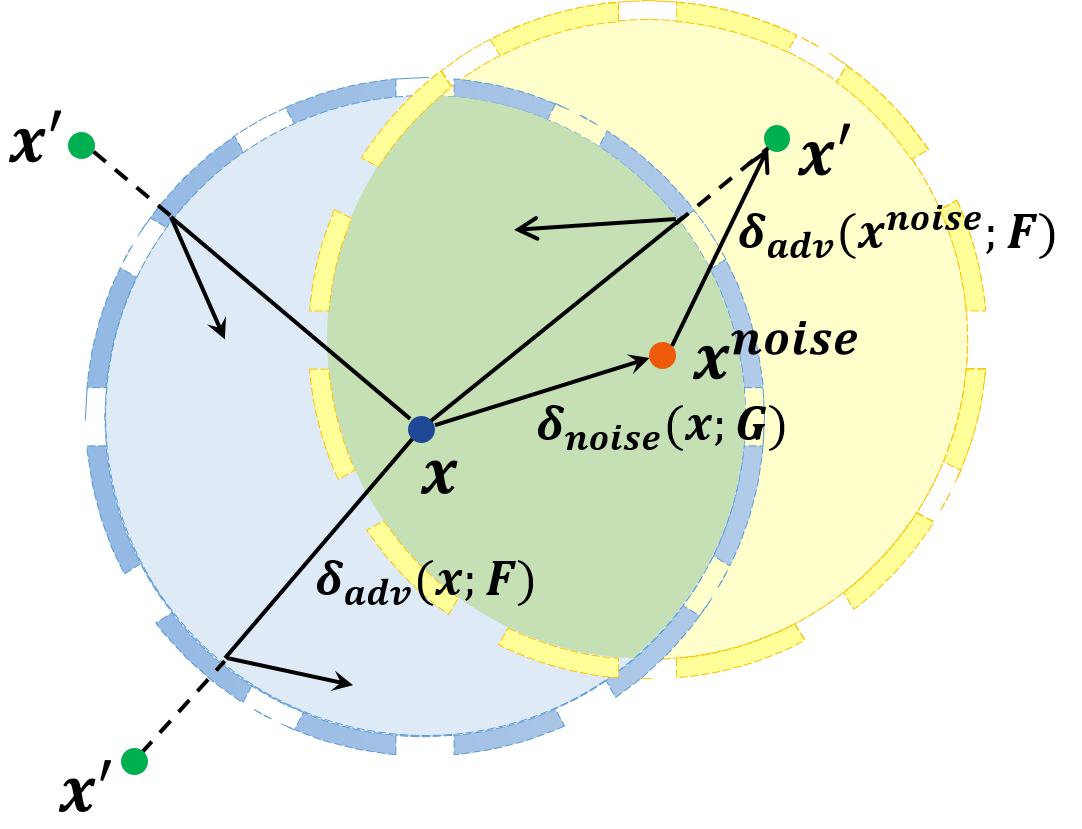}
	\caption{\textbf{Our Region-based Attack:} We simplify the entire hyperspace into a two-dimensional plane. The point $x$ stands for original image, $x'$ is our target adversarial example. For any image, it is located in a specific region corresponding to its label, and as shown in this figure, it is surrounded by its classification boundary. Attackers aim to move $x$ to the position of $x'$ out of this region by adding adversarial perturbation $\delta_{\text{adv}}(x,F)$ ($F$ is the $\epsilon$-neighborhood attack). However, distillation makes the boundary become very firm, like an ``iron wall", and points trying to rush out of the border will be ``bounced", but it can move freely inside its own region. In our region-based attack, we first let $x$ move to the position of $x^{\text{noise}}$ by adding perturbation $\delta_{\text{noise}}(x,G)$ ($G$ is the method of adding noise, here we use Gaussian noise). For the point $x^{\text{noise}}$, it has its own unbreakable boundary. Then we easily move $x^{\text{noise}}$ inside its region (adding $\delta_{\text{adv}}(x^{\text{noise}},F)$) until it arrives at the area not belonging to the limited range of $x$.}
	\label{fig:region-based-attack-idea}
\end{figure}

For the corresponding point of a fixed image, it is like its classification boundary becoming a barrier due to the effect of distillation. So it is difficult for us to add adversarial perturbations to move the point away from its original class region, as if the point is trapped in a cage with the limited range of motion. The role of the added noise is dragging the point towards the edge of the range, and the noise-add image corresponds to a new point. For the new point, it has a new range with a different barrier from the old one, and it can only move in the new region. It is nice that there is a part of region which is in the new range but not in the original range. So we can easily move it across its firm barriers (classification boundary) caused by distillation with the method of dragging it a little away from its original position at first by adding noise. Our idea is illustrated in Fig.\ref{fig:region-based-attack-idea}.

As mentioned above, the ``hard" output of distilled network leads to the initial loss value is $\infty$. So in the actual implementation, we add a little disturbance $\Delta_{f}$ for each output probability $f(x)_{i}$ to avoid meaningless value like $\log 0$. This trick can make our algorithm effectively reduce the loss to a negative value.

In addition, the adversarial examples constructed by this attack are very robust against the defense that preprocesses input images by adding noise. In other words, not only the adversarial example itself can force the network to classify it into the target class, but also after adding some noise it can still be classified into that class. The reason is that we transform an optimization problem based on a single point into one based on a region. Therefore, the final adversarial example we produce is also suitable for many images around the original one.

\subsection{Our Bypass Attack on Black-box Distilled Network with Great T Value}
The distillation temperature $T$ is an important hyper-parameter during the defensive training. If it is too small, the defense will not work effectively, but if too large, the classification accuracy of the network on the original images will be decreased. So it is neccessary for defenders to find an appropriate $T$, and attackers do not know how much it is. Our experiments show that carrying out a black-box attack on a network with a great $T$ value is quite difficult. The attack described in Section \ref{Our-Region-based-Attack} is really effective when the temperature is not too large, but its success rate is not very high for great $T$. So here we propose a bypass attack on black-box distilled network with a great $T$ value.

Our bypass attack is a circuitous mechanism. We discover that adversarial examples for different distillation temperatures have the property of transferability. So in our bypass attack, we do not confront the network directly, but choose to attack a network with a relatively low temperature $T$. Our following experiments will show that the adversarial examples generated from small-$T$ network can also attack great-$T$ network at a relatively high success rate, which is higher than attacking directly.

\section{Experimental Evaluation}
In this section, we illustrate the effectiveness of our attacks: the $\epsilon$-neighborhood attack on white-box defensively distilled networks, the region-based attack on black-box distilled model and the bypass attack on models with high distillation temperature. We will evaluate them from several aspects: success rate, runtime and image distortion (visual quality).

\subsection{Setup}

\subsubsection{Dataset}
We evaluate our attacks on three image datasets, MNIST \cite{lecun1998mnist}, CIFAR10 \cite{krizhevsky2009learning} and ImageNet \cite{deng2009imagenet}. In each dataset, we select 1000 images to test the attacks. For every image in MNIST, we generate targeted adversarial examples of all the other 9 classes, so finally we get 9000 adversarial examples. And since CIFAR10 has 10 classes too, we also make attacks on all the other 9 classes. This evaluation can fully reflect the effectiveness of our attacks.  As for ImageNet with 1000 classes itself, we randomly select any other 10 classes as attack targets. 

\subsubsection{DNN models}
For MNIST and CIFAR10 datasets, we use the same DNN models as in the C\&W attack, as shown in Table \ref{table:mnist-cifar-architecture}. And our training method and the selected hyperparameters are also the same as it. To explore the relation between our attacks and distillation temperature, we train 12 models with $T=\{1,5,10,20,30,...,90,100\}$ on each datasets. As for the huge and more challenging ImageNet dataset, we use the pretrained  Inception-v3 model \cite{szegedy2016rethinking} with 96.4\% top-5 accuracy.

\begin{table}
	\caption{Network Architectures of \textbf{MNIST} and \textbf{CIFAR10}.} 
	\label{table:mnist-cifar-architecture}
	\centering 
	\begin{tabular}{p{3cm}p{1.6cm}p{1.6cm}}
		\hline
		Layer Type&MNIST&CIFAR10\\
		\hline
		Conv.ReLU&3$\times$3$\times$32&3$\times$3$\times$64\\
		Conv.ReLU&3$\times$3$\times$32&3$\times$3$\times$64\\
		Max pooling&2$\times$2&2$\times$2\\
		Conv.ReLU&3$\times$3$\times$64&3$\times$3$\times$128\\
		Conv.ReLU&3$\times$3$\times$64&3$\times$3$\times$128\\
		Max pooling&2$\times$2&2$\times$2\\
		Fully Connect.ReLU&200&256\\
		Fully Connect.ReLU&200&256\\
		Softmax&10&10\\
		\hline
	\end{tabular}
\end{table}

\subsubsection{Implementation}
our implementation is completed on a machine with 64GB RAM, Intel Core i7-5960X CPU and two Nvidia GeForce GTX 1080 GPU cards. And all the code is based on the deep learning framework Tensorflow.

\subsection{Performance of $\epsilon$-neighborhood Attack}
\label{Performance-of-neighborhood-Attack}

\subsubsection{Success Rate and Distortion}
\begin{figure}
	\centering  
	\includegraphics[height=2.7in, width=3.6in]{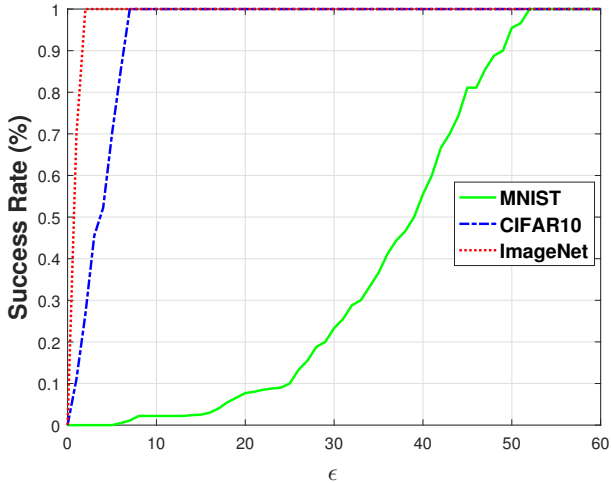}
	\caption{\textbf{Success Rate of Attack on Undistilled Model under Different Preset Maximum Perturbation $\mathbf{\epsilon}$.} For each dataset, when $\epsilon$ exceeds a certain value, the success rate can reach up to 100\%. }
	\label{Fig:epsilon-sr}
\end{figure}

\begin{figure}
	\centering  
	\includegraphics[height=2.7in, width=3.6in]{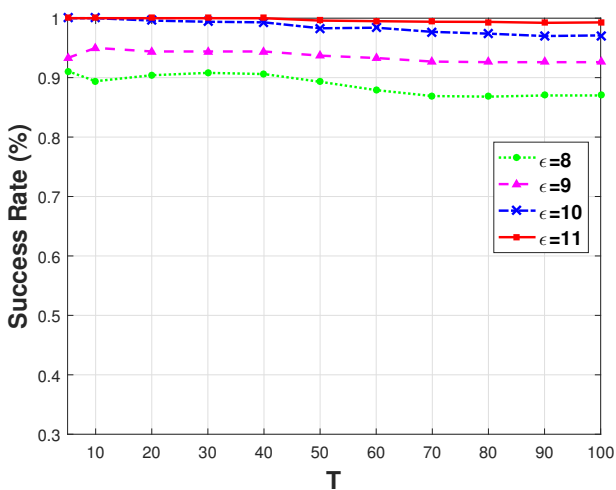}
	\caption{\textbf{Success Rate of Attack on Models with Different Distillation Temperatures on CIFAR10.} We set $\epsilon$ as \{8,9,10,11\}. Their success rates are all high, and when $\epsilon>10$, the success rate can reach up to 100\%.}
	\label{Fig:epsilon-ditisll-sr}
\end{figure}

Our $\epsilon$-neighborhood targeted attack can achieve 100\% success rate on any target label and the distortion to images is acceptable with a comparable level of C\&W attack. Better still, for RGB images in CIFAR10 and ImageNet, the visual quality of our adversarial examples is more advantageous. First we test our method on the original models for MNIST, CIFAR10 and ImageNet, Fig.\ref{Fig:epsilon-sr} shows that the attack success rate of any target class can achieve 100\% with a proper preset maximum perturbation parameter $\epsilon$, and our attack can succeed, no matter which class the original image belongs to, and which class is the target. Fig.\ref{Fig:epsilon-sr} also indicates the tradeoff between maximum perturbation and success rate. For MNIST and CIFAR10, when $\epsilon$ is greater or equal to 50 and 7 respectively, the success rate reaches up to 100\%. And for ImageNet, the $\epsilon$ value can be just 2. Actually, our experiment shows that we can achieve more than 70\% success rate even if $\epsilon$ is fixed to 1.  When applied to distilled models, our attack can also achieve almost 100\% success rate. The result on CIFAR10 can be found in Fig.\ref{Fig:epsilon-ditisll-sr}. When $\epsilon>9$, the success rate has been nearly 100\% for a large range of distillation temperature (for undistilled models, this value of $\epsilon$ is 7). So we do not need to add much more perturbation for distilled networks compared with undistilled models. 

\begin{figure}
	\begin{minipage}[t]{0.15\textwidth}
		\includegraphics[height=2.9cm,width=2.9cm]{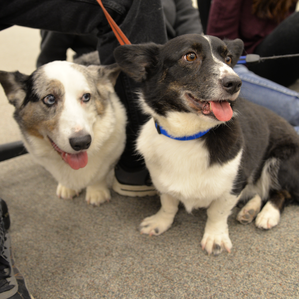}
		\caption*{(a) Original.}
	\end{minipage} 
	\hfill  
	\begin{minipage}[t]{0.15\textwidth}
		\includegraphics[height=2.9cm,width=2.9cm]{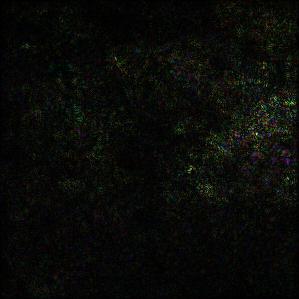}  
		\caption*{(b) C\&W.}
	\end{minipage} 
	\hfill  
	\begin{minipage}[t]{0.15\textwidth}
		\includegraphics[height=2.9cm,width=2.9cm]{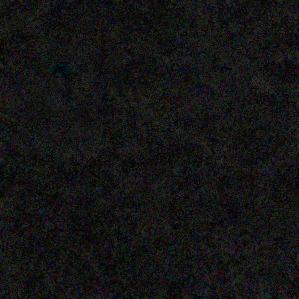}  
		\caption*{(c) Our $\epsilon$.}
	\end{minipage}
\caption{\textbf{Comparison of Adversarial Perturbation ($\mathbf{\times 30}$) between C\&W Attack and $\mathbf{\epsilon}$-neighborhood Attack.} There exist obvious bright spots in (b) C\&W attack, which indicates it modifies a lot to several pixels. But in (c) our attack, the perturbation is more imperceptible.}
\label{Fig:cw-our}
\end{figure}

\begin{figure}[t]  
		\centerline{Target Class}
		\centerline{0\hspace{0.72cm}1\hspace{0.72cm}2\hspace{0.72cm}3\hspace{0.72cm}4\hspace{0.72cm}5\hspace{0.72cm}6\hspace{0.72cm}7\hspace{0.72cm}8\hspace{0.72cm}9}
		\vfill
		\vspace{3pt}
		\begin{minipage}[t]{0.043\textwidth}
			\underline{\includegraphics[height=0.8cm,width=0.8cm]{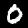}}
		\end{minipage} 
		\hfill  
		\begin{minipage}[t]{0.043\textwidth}
			\includegraphics[height=0.8cm,width=0.8cm]{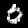}  
		\end{minipage} 
		\hfill  
		\begin{minipage}[t]{0.043\textwidth}
			\includegraphics[height=0.8cm,width=0.8cm]{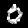}  
		\end{minipage} 
		\hfill  
		\begin{minipage}[t]{0.043\textwidth}
			\includegraphics[height=0.8cm,width=0.8cm]{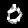}  
		\end{minipage} 
		\hfill  
		\begin{minipage}[t]{0.043\textwidth}
			\includegraphics[height=0.8cm,width=0.8cm]{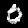}  
		\end{minipage} 
		\hfill  
		\begin{minipage}[t]{0.043\textwidth}
			\includegraphics[height=0.8cm,width=0.8cm]{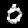}  
		\end{minipage} 
		\hfill  
		\begin{minipage}[t]{0.043\textwidth}
			\includegraphics[height=0.8cm,width=0.8cm]{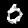}  
		\end{minipage} 
		\hfill  
		\begin{minipage}[t]{0.043\textwidth}
			\includegraphics[height=0.8cm,width=0.8cm]{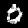}  
		\end{minipage} 
		\hfill  
		\begin{minipage}[t]{0.043\textwidth}
			\includegraphics[height=0.8cm,width=0.8cm]{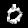}  
		\end{minipage} 
		\hfill  
		\begin{minipage}[t]{0.043\textwidth}
			\includegraphics[height=0.8cm,width=0.8cm]{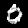} 
		\end{minipage} 
		\vfill
		\vspace{1pt}
		\begin{minipage}[t]{0.043\textwidth}
			\includegraphics[height=0.8cm,width=0.8cm]{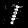}  
		\end{minipage} 
		\hfill
		\begin{minipage}[t]{0.043\textwidth}
			\underline{\includegraphics[height=0.8cm,width=0.8cm]{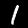}}  
		\end{minipage} 
		\hfill
		\begin{minipage}[t]{0.043\textwidth}
			\includegraphics[height=0.8cm,width=0.8cm]{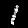}  
		\end{minipage} 
		\hfill
		\begin{minipage}[t]{0.043\textwidth}
			\includegraphics[height=0.8cm,width=0.8cm]{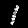}  
		\end{minipage} 
		\hfill
		\begin{minipage}[t]{0.043\textwidth}
			\includegraphics[height=0.8cm,width=0.8cm]{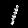}  
		\end{minipage} 
		\hfill
		\begin{minipage}[t]{0.043\textwidth}
			\includegraphics[height=0.8cm,width=0.8cm]{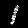}  
		\end{minipage} 
		\hfill
		\begin{minipage}[t]{0.043\textwidth}
			\includegraphics[height=0.8cm,width=0.8cm]{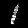}  
		\end{minipage} 
		\hfill
		\begin{minipage}[t]{0.043\textwidth}
			\includegraphics[height=0.8cm,width=0.8cm]{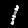}  
		\end{minipage} 
		\hfill
		\begin{minipage}[t]{0.043\textwidth}
			\includegraphics[height=0.8cm,width=0.8cm]{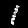}  
		\end{minipage} 
		\hfill
		\begin{minipage}[t]{0.043\textwidth}
			\includegraphics[height=0.8cm,width=0.8cm]{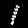}  
		\end{minipage} 
		\vfill
		\vspace{1pt}
		\begin{minipage}[t]{0.043\textwidth}
			\includegraphics[height=0.8cm,width=0.8cm]{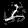}  
		\end{minipage} 
		\hfill
		\begin{minipage}[t]{0.043\textwidth}
			\includegraphics[height=0.8cm,width=0.8cm]{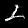}  
		\end{minipage} 
		\hfill
		\begin{minipage}[t]{0.043\textwidth}
			\underline{\includegraphics[height=0.8cm,width=0.8cm]{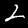}}  
		\end{minipage} 
		\hfill
		\begin{minipage}[t]{0.043\textwidth}
			\includegraphics[height=0.8cm,width=0.8cm]{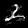}  
		\end{minipage} 
		\hfill
		\begin{minipage}[t]{0.043\textwidth}
			\includegraphics[height=0.8cm,width=0.8cm]{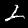}  
		\end{minipage} 
		\hfill
		\begin{minipage}[t]{0.043\textwidth}
			\includegraphics[height=0.8cm,width=0.8cm]{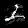}  
		\end{minipage} 
		\hfill
		\begin{minipage}[t]{0.043\textwidth}
			\includegraphics[height=0.8cm,width=0.8cm]{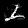}  
		\end{minipage} 
		\hfill
		\begin{minipage}[t]{0.043\textwidth}
			\includegraphics[height=0.8cm,width=0.8cm]{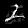}  
		\end{minipage} 
		\hfill
		\begin{minipage}[t]{0.043\textwidth}
			\includegraphics[height=0.8cm,width=0.8cm]{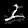}  
		\end{minipage} 
		\hfill
		\begin{minipage}[t]{0.043\textwidth}
			\includegraphics[height=0.8cm,width=0.8cm]{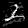}  
		\end{minipage} 
		\vfill
		\vspace{1pt}
		\begin{minipage}[t]{0.043\textwidth}
			\includegraphics[height=0.8cm,width=0.8cm]{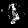}  
		\end{minipage} 
		\hfill
		\begin{minipage}[t]{0.043\textwidth}
			\includegraphics[height=0.8cm,width=0.8cm]{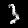}  
		\end{minipage} 
		\hfill
		\begin{minipage}[t]{0.043\textwidth}
			\includegraphics[height=0.8cm,width=0.8cm]{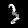}  
		\end{minipage} 
		\hfill
		\begin{minipage}[t]{0.043\textwidth}
			\underline{\includegraphics[height=0.8cm,width=0.8cm]{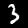}}  
		\end{minipage} 
		\hfill
		\begin{minipage}[t]{0.043\textwidth}
			\includegraphics[height=0.8cm,width=0.8cm]{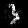}  
		\end{minipage} 
		\hfill
		\begin{minipage}[t]{0.043\textwidth}
			\includegraphics[height=0.8cm,width=0.8cm]{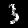}  
		\end{minipage} 
		\hfill
		\begin{minipage}[t]{0.043\textwidth}
			\includegraphics[height=0.8cm,width=0.8cm]{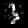}  
		\end{minipage} 
		\hfill
		\begin{minipage}[t]{0.043\textwidth}
			\includegraphics[height=0.8cm,width=0.8cm]{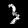}  
		\end{minipage} 
		\hfill
		\begin{minipage}[t]{0.043\textwidth}
			\includegraphics[height=0.8cm,width=0.8cm]{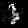}  
		\end{minipage} 
		\hfill
		\begin{minipage}[t]{0.043\textwidth}
			\includegraphics[height=0.8cm,width=0.8cm]{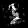}  
		\end{minipage} 
		\vfill
		\vspace{1pt}
		\begin{minipage}[t]{0.043\textwidth}
			\includegraphics[height=0.8cm,width=0.8cm]{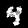}  
		\end{minipage} 
		\hfill
		\begin{minipage}[t]{0.043\textwidth}
			\includegraphics[height=0.8cm,width=0.8cm]{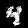}  
		\end{minipage} 
		\hfill
		\begin{minipage}[t]{0.043\textwidth}
			\includegraphics[height=0.8cm,width=0.8cm]{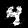}  
		\end{minipage} 
		\hfill
		\begin{minipage}[t]{0.043\textwidth}
			\includegraphics[height=0.8cm,width=0.8cm]{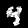}  
		\end{minipage} 
		\hfill
		\begin{minipage}[t]{0.043\textwidth}
			\underline{\includegraphics[height=0.8cm,width=0.8cm]{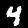}}  
		\end{minipage} 
		\hfill
		\begin{minipage}[t]{0.043\textwidth}
			\includegraphics[height=0.8cm,width=0.8cm]{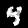}  
		\end{minipage} 
		\hfill
		\begin{minipage}[t]{0.043\textwidth}
			\includegraphics[height=0.8cm,width=0.8cm]{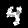}  
		\end{minipage} 
		\hfill
		\begin{minipage}[t]{0.043\textwidth}
			\includegraphics[height=0.8cm,width=0.8cm]{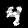}  
		\end{minipage} 
		\hfill
		\begin{minipage}[t]{0.043\textwidth}
			\includegraphics[height=0.8cm,width=0.8cm]{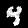}  
		\end{minipage} 
		\hfill
		\begin{minipage}[t]{0.043\textwidth}
			\includegraphics[height=0.8cm,width=0.8cm]{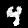}  
		\end{minipage} 
		\vfill
		\vspace{1pt}
		\begin{minipage}[t]{0.043\textwidth}
			\includegraphics[height=0.8cm,width=0.8cm]{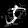}  
		\end{minipage} 
		\hfill
		\begin{minipage}[t]{0.043\textwidth}
			\includegraphics[height=0.8cm,width=0.8cm]{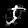}  
		\end{minipage} 
		\hfill
		\begin{minipage}[t]{0.043\textwidth}
			\includegraphics[height=0.8cm,width=0.8cm]{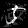}  
		\end{minipage} 
		\hfill
		\begin{minipage}[t]{0.043\textwidth}
			\includegraphics[height=0.8cm,width=0.8cm]{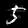}  
		\end{minipage} 
		\hfill
		\begin{minipage}[t]{0.043\textwidth}
			\includegraphics[height=0.8cm,width=0.8cm]{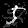}  
		\end{minipage} 
		\hfill
		\begin{minipage}[t]{0.043\textwidth}
			\underline{\includegraphics[height=0.8cm,width=0.8cm]{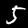}} 
		\end{minipage} 
		\hfill
		\begin{minipage}[t]{0.043\textwidth}
			\includegraphics[height=0.8cm,width=0.8cm]{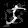}  
		\end{minipage} 
		\hfill
		\begin{minipage}[t]{0.043\textwidth}
			\includegraphics[height=0.8cm,width=0.8cm]{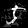}  
		\end{minipage} 
		\hfill
		\begin{minipage}[t]{0.043\textwidth}
			\includegraphics[height=0.8cm,width=0.8cm]{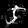}  
		\end{minipage} 
		\hfill
		\begin{minipage}[t]{0.043\textwidth}
			\includegraphics[height=0.8cm,width=0.8cm]{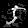}  
		\end{minipage} 
		\vfill
		\vspace{1pt}
		\begin{minipage}[t]{0.043\textwidth}
			\includegraphics[height=0.8cm,width=0.8cm]{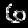}  
		\end{minipage} 
		\hfill
		\begin{minipage}[t]{0.043\textwidth}
			\includegraphics[height=0.8cm,width=0.8cm]{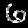}  
		\end{minipage} 
		\hfill
		\begin{minipage}[t]{0.043\textwidth}
			\includegraphics[height=0.8cm,width=0.8cm]{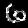}  
		\end{minipage} 
		\hfill
		\begin{minipage}[t]{0.043\textwidth}
			\includegraphics[height=0.8cm,width=0.8cm]{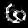}  
		\end{minipage} 
		\hfill
		\begin{minipage}[t]{0.043\textwidth}
			\includegraphics[height=0.8cm,width=0.8cm]{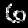}  
		\end{minipage} 
		\hfill
		\begin{minipage}[t]{0.043\textwidth}
			\includegraphics[height=0.8cm,width=0.8cm]{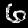}  
		\end{minipage} 
		\hfill
		\begin{minipage}[t]{0.043\textwidth}
			\underline{\includegraphics[height=0.8cm,width=0.8cm]{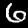}}  
		\end{minipage} 
		\hfill
		\begin{minipage}[t]{0.043\textwidth}
			\includegraphics[height=0.8cm,width=0.8cm]{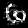}  
		\end{minipage} 
		\hfill
		\begin{minipage}[t]{0.043\textwidth}
			\includegraphics[height=0.8cm,width=0.8cm]{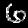}  
		\end{minipage} 
		\hfill
		\begin{minipage}[t]{0.043\textwidth}
			\includegraphics[height=0.8cm,width=0.8cm]{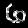}  
		\end{minipage} 
		\vfill
		\vspace{1pt}
		\begin{minipage}[t]{0.043\textwidth}
			\includegraphics[height=0.8cm,width=0.8cm]{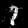}  
		\end{minipage} 
		\hfill
		\begin{minipage}[t]{0.043\textwidth}
			\includegraphics[height=0.8cm,width=0.8cm]{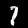}  
		\end{minipage} 
		\hfill
		\begin{minipage}[t]{0.043\textwidth}
			\includegraphics[height=0.8cm,width=0.8cm]{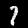}  
		\end{minipage} 
		\hfill
		\begin{minipage}[t]{0.043\textwidth}
			\includegraphics[height=0.8cm,width=0.8cm]{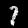}  
		\end{minipage} 
		\hfill
		\begin{minipage}[t]{0.043\textwidth}
			\includegraphics[height=0.8cm,width=0.8cm]{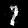}  
		\end{minipage} 
		\hfill
		\begin{minipage}[t]{0.043\textwidth}
			\includegraphics[height=0.8cm,width=0.8cm]{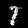}  
		\end{minipage} 
		\hfill
		\begin{minipage}[t]{0.043\textwidth}
			\includegraphics[height=0.8cm,width=0.8cm]{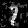}  
		\end{minipage} 
		\hfill
		\begin{minipage}[t]{0.043\textwidth}
			\underline{\includegraphics[height=0.8cm,width=0.8cm]{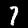}}
		\end{minipage} 
		\hfill
		\begin{minipage}[t]{0.043\textwidth}
			\includegraphics[height=0.8cm,width=0.8cm]{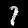}  
		\end{minipage} 
		\hfill
		\begin{minipage}[t]{0.043\textwidth}
			\includegraphics[height=0.8cm,width=0.8cm]{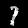}  
		\end{minipage} 
		\vfill
		\vspace{1pt}
		\begin{minipage}[t]{0.043\textwidth}
			\includegraphics[height=0.8cm,width=0.8cm]{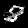}  
		\end{minipage} 
		\hfill
		\begin{minipage}[t]{0.043\textwidth}
			\includegraphics[height=0.8cm,width=0.8cm]{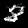}  
		\end{minipage} 
		\hfill
		\begin{minipage}[t]{0.043\textwidth}
			\includegraphics[height=0.8cm,width=0.8cm]{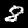}  
		\end{minipage} 
		\hfill
		\begin{minipage}[t]{0.043\textwidth}
			\includegraphics[height=0.8cm,width=0.8cm]{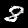}  
		\end{minipage} 
		\hfill
		\begin{minipage}[t]{0.043\textwidth}
			\includegraphics[height=0.8cm,width=0.8cm]{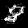}  
		\end{minipage} 
		\hfill
		\begin{minipage}[t]{0.043\textwidth}
			\includegraphics[height=0.8cm,width=0.8cm]{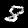}  
		\end{minipage} 
		\hfill
		\begin{minipage}[t]{0.043\textwidth}
			\includegraphics[height=0.8cm,width=0.8cm]{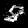}  
		\end{minipage} 
		\hfill
		\begin{minipage}[t]{0.043\textwidth}
			\includegraphics[height=0.8cm,width=0.8cm]{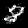}  
		\end{minipage} 
		\hfill
		\begin{minipage}[t]{0.043\textwidth}
			\underline{\includegraphics[height=0.8cm,width=0.8cm]{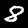}}  
		\end{minipage} 
		\hfill
		\begin{minipage}[t]{0.043\textwidth}
			\includegraphics[height=0.8cm,width=0.8cm]{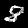}  
		\end{minipage} 
		\vfill
		\vspace{1pt}
		\begin{minipage}[t]{0.043\textwidth}
			\includegraphics[height=0.8cm,width=0.8cm]{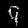}  
		\end{minipage} 
		\hfill
		\begin{minipage}[t]{0.043\textwidth}
			\includegraphics[height=0.8cm,width=0.8cm]{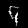}  
		\end{minipage} 
		\hfill
		\begin{minipage}[t]{0.043\textwidth}
			\includegraphics[height=0.8cm,width=0.8cm]{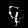}  
		\end{minipage} 
		\hfill
		\begin{minipage}[t]{0.043\textwidth}
			\includegraphics[height=0.8cm,width=0.8cm]{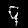}  
		\end{minipage} 
		\hfill
		\begin{minipage}[t]{0.043\textwidth}
			\includegraphics[height=0.8cm,width=0.8cm]{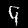}  
		\end{minipage} 
		\hfill
		\begin{minipage}[t]{0.043\textwidth}
			\includegraphics[height=0.8cm,width=0.8cm]{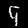}  
		\end{minipage} 
		\hfill
		\begin{minipage}[t]{0.043\textwidth}
			\includegraphics[height=0.8cm,width=0.8cm]{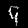}  
		\end{minipage} 
		\hfill
		\begin{minipage}[t]{0.043\textwidth}
			\includegraphics[height=0.8cm,width=0.8cm]{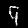}  
		\end{minipage} 
		\hfill
		\begin{minipage}[t]{0.043\textwidth}
			\includegraphics[height=0.8cm,width=0.8cm]{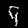}  
		\end{minipage} 
		\hfill
		\begin{minipage}[t]{0.043\textwidth}
			\underline{\includegraphics[height=0.8cm,width=0.8cm]{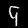}}  
		\end{minipage} 
		
		\caption{\textbf{Targeted Adversarial Examples on MNIST}. We randomly selected 10 images, respectively belonging to class \{0,1,2,...,9\}. For each test image, we generate adversarial examples of all the other 9 target classes, as shown in each row. The underlined image on the diagonal is the original image.}
		\label{Fig:mnist-show}
	\end{figure}

\begin{figure}[t]  
	\centerline{Target Class}
	\centerline{1\hspace{0.72cm}2\hspace{0.72cm}3\hspace{0.72cm}4\hspace{0.72cm}5\hspace{0.72cm}6\hspace{0.72cm}7\hspace{0.72cm}8\hspace{0.72cm}9\hspace{0.72cm}10}
	\vfill
	\vspace{3pt}
	\begin{minipage}[t]{0.043\textwidth}
		\underline{\includegraphics[height=0.8cm,width=0.8cm]{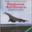}}
	\end{minipage} 
	\hfill  
	\begin{minipage}[t]{0.043\textwidth}
		\includegraphics[height=0.8cm,width=0.8cm]{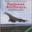}  
	\end{minipage} 
	\hfill  
    \begin{minipage}[t]{0.043\textwidth}
	\includegraphics[height=0.8cm,width=0.8cm]{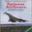}  
    \end{minipage} 
    \hfill  
    \begin{minipage}[t]{0.043\textwidth}
	\includegraphics[height=0.8cm,width=0.8cm]{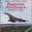}  
    \end{minipage} 
    \hfill  
    \begin{minipage}[t]{0.043\textwidth}
	\includegraphics[height=0.8cm,width=0.8cm]{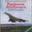}  
    \end{minipage} 
    \hfill  
    \begin{minipage}[t]{0.043\textwidth}
	\includegraphics[height=0.8cm,width=0.8cm]{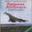}  
    \end{minipage} 
    \hfill  
    \begin{minipage}[t]{0.043\textwidth}
	\includegraphics[height=0.8cm,width=0.8cm]{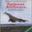}  
    \end{minipage} 
    \hfill  
    \begin{minipage}[t]{0.043\textwidth}
	\includegraphics[height=0.8cm,width=0.8cm]{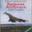}  
    \end{minipage} 
    \hfill  
    \begin{minipage}[t]{0.043\textwidth}
	\includegraphics[height=0.8cm,width=0.8cm]{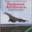}  
    \end{minipage} 
    \hfill  
    \begin{minipage}[t]{0.043\textwidth}
	\includegraphics[height=0.8cm,width=0.8cm]{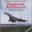} 
    \end{minipage} 
	\vfill
	\vspace{1pt}
	\begin{minipage}[t]{0.043\textwidth}
		\includegraphics[height=0.8cm,width=0.8cm]{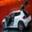}  
	\end{minipage} 
	\hfill
	\begin{minipage}[t]{0.043\textwidth}
		\underline{\includegraphics[height=0.8cm,width=0.8cm]{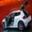}}  
	\end{minipage} 
	\hfill
	\begin{minipage}[t]{0.043\textwidth}
		\includegraphics[height=0.8cm,width=0.8cm]{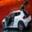}  
	\end{minipage} 
	\hfill
	\begin{minipage}[t]{0.043\textwidth}
		\includegraphics[height=0.8cm,width=0.8cm]{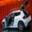}  
	\end{minipage} 
	\hfill
	\begin{minipage}[t]{0.043\textwidth}
		\includegraphics[height=0.8cm,width=0.8cm]{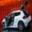}  
	\end{minipage} 
	\hfill
	\begin{minipage}[t]{0.043\textwidth}
		\includegraphics[height=0.8cm,width=0.8cm]{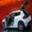}  
	\end{minipage} 
	\hfill
	\begin{minipage}[t]{0.043\textwidth}
		\includegraphics[height=0.8cm,width=0.8cm]{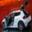}  
	\end{minipage} 
	\hfill
	\begin{minipage}[t]{0.043\textwidth}
		\includegraphics[height=0.8cm,width=0.8cm]{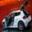}  
	\end{minipage} 
	\hfill
	\begin{minipage}[t]{0.043\textwidth}
		\includegraphics[height=0.8cm,width=0.8cm]{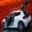}  
	\end{minipage} 
	\hfill
	\begin{minipage}[t]{0.043\textwidth}
		\includegraphics[height=0.8cm,width=0.8cm]{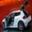}  
	\end{minipage} 
	\vfill
	\vspace{1pt}
	\begin{minipage}[t]{0.043\textwidth}
		\includegraphics[height=0.8cm,width=0.8cm]{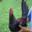}  
	\end{minipage} 
	\hfill
	\begin{minipage}[t]{0.043\textwidth}
		\includegraphics[height=0.8cm,width=0.8cm]{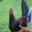}  
	\end{minipage} 
	\hfill
	\begin{minipage}[t]{0.043\textwidth}
		\underline{\includegraphics[height=0.8cm,width=0.8cm]{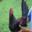}}  
	\end{minipage} 
	\hfill
	\begin{minipage}[t]{0.043\textwidth}
		\includegraphics[height=0.8cm,width=0.8cm]{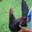}  
	\end{minipage} 
	\hfill
	\begin{minipage}[t]{0.043\textwidth}
		\includegraphics[height=0.8cm,width=0.8cm]{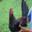}  
	\end{minipage} 
	\hfill
	\begin{minipage}[t]{0.043\textwidth}
		\includegraphics[height=0.8cm,width=0.8cm]{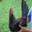}  
	\end{minipage} 
	\hfill
	\begin{minipage}[t]{0.043\textwidth}
		\includegraphics[height=0.8cm,width=0.8cm]{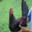}  
	\end{minipage} 
	\hfill
	\begin{minipage}[t]{0.043\textwidth}
		\includegraphics[height=0.8cm,width=0.8cm]{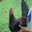}  
	\end{minipage} 
	\hfill
	\begin{minipage}[t]{0.043\textwidth}
		\includegraphics[height=0.8cm,width=0.8cm]{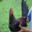}  
	\end{minipage} 
	\hfill
	\begin{minipage}[t]{0.043\textwidth}
		\includegraphics[height=0.8cm,width=0.8cm]{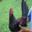}  
	\end{minipage} 
\vfill
\vspace{1pt}
	\begin{minipage}[t]{0.043\textwidth}
	\includegraphics[height=0.8cm,width=0.8cm]{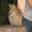}  
\end{minipage} 
\hfill
\begin{minipage}[t]{0.043\textwidth}
	\includegraphics[height=0.8cm,width=0.8cm]{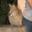}  
\end{minipage} 
\hfill
\begin{minipage}[t]{0.043\textwidth}
	\includegraphics[height=0.8cm,width=0.8cm]{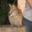}  
\end{minipage} 
\hfill
\begin{minipage}[t]{0.043\textwidth}
	\underline{\includegraphics[height=0.8cm,width=0.8cm]{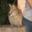}}  
\end{minipage} 
\hfill
\begin{minipage}[t]{0.043\textwidth}
	\includegraphics[height=0.8cm,width=0.8cm]{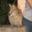}  
\end{minipage} 
\hfill
\begin{minipage}[t]{0.043\textwidth}
	\includegraphics[height=0.8cm,width=0.8cm]{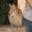}  
\end{minipage} 
\hfill
\begin{minipage}[t]{0.043\textwidth}
	\includegraphics[height=0.8cm,width=0.8cm]{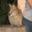}  
\end{minipage} 
\hfill
\begin{minipage}[t]{0.043\textwidth}
	\includegraphics[height=0.8cm,width=0.8cm]{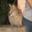}  
\end{minipage} 
\hfill
\begin{minipage}[t]{0.043\textwidth}
	\includegraphics[height=0.8cm,width=0.8cm]{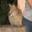}  
\end{minipage} 
\hfill
\begin{minipage}[t]{0.043\textwidth}
	\includegraphics[height=0.8cm,width=0.8cm]{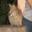}  
\end{minipage} 
\vfill
\vspace{1pt}
	\begin{minipage}[t]{0.043\textwidth}
	\includegraphics[height=0.8cm,width=0.8cm]{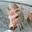}  
\end{minipage} 
\hfill
\begin{minipage}[t]{0.043\textwidth}
	\includegraphics[height=0.8cm,width=0.8cm]{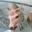}  
\end{minipage} 
\hfill
\begin{minipage}[t]{0.043\textwidth}
	\includegraphics[height=0.8cm,width=0.8cm]{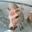}  
\end{minipage} 
\hfill
\begin{minipage}[t]{0.043\textwidth}
	\includegraphics[height=0.8cm,width=0.8cm]{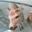}  
\end{minipage} 
\hfill
\begin{minipage}[t]{0.043\textwidth}
	\underline{\includegraphics[height=0.8cm,width=0.8cm]{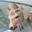}}  
\end{minipage} 
\hfill
\begin{minipage}[t]{0.043\textwidth}
	\includegraphics[height=0.8cm,width=0.8cm]{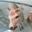}  
\end{minipage} 
\hfill
\begin{minipage}[t]{0.043\textwidth}
	\includegraphics[height=0.8cm,width=0.8cm]{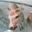}  
\end{minipage} 
\hfill
\begin{minipage}[t]{0.043\textwidth}
	\includegraphics[height=0.8cm,width=0.8cm]{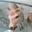}  
\end{minipage} 
\hfill
\begin{minipage}[t]{0.043\textwidth}
	\includegraphics[height=0.8cm,width=0.8cm]{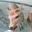}  
\end{minipage} 
\hfill
\begin{minipage}[t]{0.043\textwidth}
	\includegraphics[height=0.8cm,width=0.8cm]{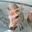}  
\end{minipage} 
\vfill
\vspace{1pt}
	\begin{minipage}[t]{0.043\textwidth}
	\includegraphics[height=0.8cm,width=0.8cm]{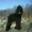}  
\end{minipage} 
\hfill
\begin{minipage}[t]{0.043\textwidth}
	\includegraphics[height=0.8cm,width=0.8cm]{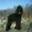}  
\end{minipage} 
\hfill
\begin{minipage}[t]{0.043\textwidth}
	\includegraphics[height=0.8cm,width=0.8cm]{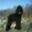}  
\end{minipage} 
\hfill
\begin{minipage}[t]{0.043\textwidth}
	\includegraphics[height=0.8cm,width=0.8cm]{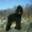}  
\end{minipage} 
\hfill
\begin{minipage}[t]{0.043\textwidth}
	\includegraphics[height=0.8cm,width=0.8cm]{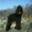}  
\end{minipage} 
\hfill
\begin{minipage}[t]{0.043\textwidth}
	\underline{\includegraphics[height=0.8cm,width=0.8cm]{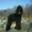}} 
\end{minipage} 
\hfill
\begin{minipage}[t]{0.043\textwidth}
	\includegraphics[height=0.8cm,width=0.8cm]{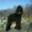}  
\end{minipage} 
\hfill
\begin{minipage}[t]{0.043\textwidth}
	\includegraphics[height=0.8cm,width=0.8cm]{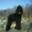}  
\end{minipage} 
\hfill
\begin{minipage}[t]{0.043\textwidth}
	\includegraphics[height=0.8cm,width=0.8cm]{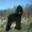}  
\end{minipage} 
\hfill
\begin{minipage}[t]{0.043\textwidth}
	\includegraphics[height=0.8cm,width=0.8cm]{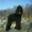}  
\end{minipage} 
\vfill
\vspace{1pt}
	\begin{minipage}[t]{0.043\textwidth}
	\includegraphics[height=0.8cm,width=0.8cm]{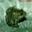}  
\end{minipage} 
\hfill
\begin{minipage}[t]{0.043\textwidth}
	\includegraphics[height=0.8cm,width=0.8cm]{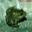}  
\end{minipage} 
\hfill
\begin{minipage}[t]{0.043\textwidth}
	\includegraphics[height=0.8cm,width=0.8cm]{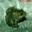}  
\end{minipage} 
\hfill
\begin{minipage}[t]{0.043\textwidth}
	\includegraphics[height=0.8cm,width=0.8cm]{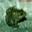}  
\end{minipage} 
\hfill
\begin{minipage}[t]{0.043\textwidth}
	\includegraphics[height=0.8cm,width=0.8cm]{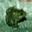}  
\end{minipage} 
\hfill
\begin{minipage}[t]{0.043\textwidth}
	\includegraphics[height=0.8cm,width=0.8cm]{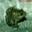}  
\end{minipage} 
\hfill
\begin{minipage}[t]{0.043\textwidth}
	\underline{\includegraphics[height=0.8cm,width=0.8cm]{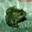}}  
\end{minipage} 
\hfill
\begin{minipage}[t]{0.043\textwidth}
	\includegraphics[height=0.8cm,width=0.8cm]{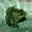}  
\end{minipage} 
\hfill
\begin{minipage}[t]{0.043\textwidth}
	\includegraphics[height=0.8cm,width=0.8cm]{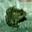}  
\end{minipage} 
\hfill
\begin{minipage}[t]{0.043\textwidth}
	\includegraphics[height=0.8cm,width=0.8cm]{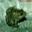}  
\end{minipage} 
\vfill
\vspace{1pt}
	\begin{minipage}[t]{0.043\textwidth}
	\includegraphics[height=0.8cm,width=0.8cm]{cifar_7_0.jpg}  
\end{minipage} 
\hfill
\begin{minipage}[t]{0.043\textwidth}
	\includegraphics[height=0.8cm,width=0.8cm]{cifar_7_1.jpg}  
\end{minipage} 
\hfill
\begin{minipage}[t]{0.043\textwidth}
	\includegraphics[height=0.8cm,width=0.8cm]{cifar_7_2.jpg}  
\end{minipage} 
\hfill
\begin{minipage}[t]{0.043\textwidth}
	\includegraphics[height=0.8cm,width=0.8cm]{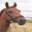}  
\end{minipage} 
\hfill
\begin{minipage}[t]{0.043\textwidth}
	\includegraphics[height=0.8cm,width=0.8cm]{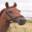}  
\end{minipage} 
\hfill
\begin{minipage}[t]{0.043\textwidth}
	\includegraphics[height=0.8cm,width=0.8cm]{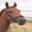}  
\end{minipage} 
\hfill
\begin{minipage}[t]{0.043\textwidth}
	\includegraphics[height=0.8cm,width=0.8cm]{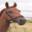}  
\end{minipage} 
\hfill
\begin{minipage}[t]{0.043\textwidth}
	\underline{\includegraphics[height=0.8cm,width=0.8cm]{cifar_7_7.jpg}}
\end{minipage} 
\hfill
\begin{minipage}[t]{0.043\textwidth}
	\includegraphics[height=0.8cm,width=0.8cm]{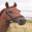}  
\end{minipage} 
\hfill
\begin{minipage}[t]{0.043\textwidth}
	\includegraphics[height=0.8cm,width=0.8cm]{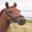}  
\end{minipage} 
\vfill
\vspace{1pt}
	\begin{minipage}[t]{0.043\textwidth}
	\includegraphics[height=0.8cm,width=0.8cm]{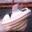}  
\end{minipage} 
\hfill
\begin{minipage}[t]{0.043\textwidth}
	\includegraphics[height=0.8cm,width=0.8cm]{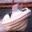}  
\end{minipage} 
\hfill
\begin{minipage}[t]{0.043\textwidth}
	\includegraphics[height=0.8cm,width=0.8cm]{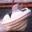}  
\end{minipage} 
\hfill
\begin{minipage}[t]{0.043\textwidth}
	\includegraphics[height=0.8cm,width=0.8cm]{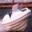}  
\end{minipage} 
\hfill
\begin{minipage}[t]{0.043\textwidth}
	\includegraphics[height=0.8cm,width=0.8cm]{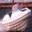}  
\end{minipage} 
\hfill
\begin{minipage}[t]{0.043\textwidth}
	\includegraphics[height=0.8cm,width=0.8cm]{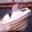}  
\end{minipage} 
\hfill
\begin{minipage}[t]{0.043\textwidth}
	\includegraphics[height=0.8cm,width=0.8cm]{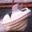}  
\end{minipage} 
\hfill
\begin{minipage}[t]{0.043\textwidth}
	\includegraphics[height=0.8cm,width=0.8cm]{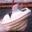}  
\end{minipage} 
\hfill
\begin{minipage}[t]{0.043\textwidth}
	\underline{\includegraphics[height=0.8cm,width=0.8cm]{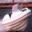}}  
\end{minipage} 
\hfill
\begin{minipage}[t]{0.043\textwidth}
	\includegraphics[height=0.8cm,width=0.8cm]{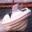}  
\end{minipage} 
\vfill
\vspace{1pt}
	\begin{minipage}[t]{0.043\textwidth}
	\includegraphics[height=0.8cm,width=0.8cm]{cifar_9_0.jpg}  
\end{minipage} 
\hfill
\begin{minipage}[t]{0.043\textwidth}
	\includegraphics[height=0.8cm,width=0.8cm]{cifar_9_1.jpg}  
\end{minipage} 
\hfill
\begin{minipage}[t]{0.043\textwidth}
	\includegraphics[height=0.8cm,width=0.8cm]{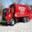}  
\end{minipage} 
\hfill
\begin{minipage}[t]{0.043\textwidth}
	\includegraphics[height=0.8cm,width=0.8cm]{cifar_9_3.jpg}  
\end{minipage} 
\hfill
\begin{minipage}[t]{0.043\textwidth}
	\includegraphics[height=0.8cm,width=0.8cm]{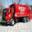}  
\end{minipage} 
\hfill
\begin{minipage}[t]{0.043\textwidth}
	\includegraphics[height=0.8cm,width=0.8cm]{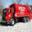}  
\end{minipage} 
\hfill
\begin{minipage}[t]{0.043\textwidth}
	\includegraphics[height=0.8cm,width=0.8cm]{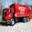}  
\end{minipage} 
\hfill
\begin{minipage}[t]{0.043\textwidth}
	\includegraphics[height=0.8cm,width=0.8cm]{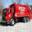}  
\end{minipage} 
\hfill
\begin{minipage}[t]{0.043\textwidth}
	\includegraphics[height=0.8cm,width=0.8cm]{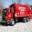}  
\end{minipage} 
\hfill
\begin{minipage}[t]{0.043\textwidth}
	\underline{\includegraphics[height=0.8cm,width=0.8cm]{cifar_9_9.jpg}}  
\end{minipage} 

	\caption{\textbf{Targeted Adversarial Examples on CIFAF10}. We randomly selected 10 images, respectively belonging to class \{1,2,...,10\}. For each test image, we generate adversarial examples of all the other 9 target classes, as shown in each row. The underlined image on the diagonal is the original image.}
	\label{Fig:cifar-show}
\end{figure}

\begin{figure*}[h]  
		\centerline{\large{Target Class}}
		\centerline{\large{original}\hspace{1.7cm}\large{1}\hspace{1.38cm}\large{2}\hspace{1.38cm}\large{3}\hspace{1.38cm}\large{4}\hspace{1.38cm}\large{5}\hspace{1.38cm}\large{6}\hspace{1.38cm}\large{7}\hspace{1.38cm}\large{8}\hspace{1.38cm}\large{9}\hspace{1.38cm}\large{10}\hspace{0.5cm}}
		\vfill
		\vspace{3pt}
		\begin{minipage}[t]{0.08\textwidth}
			\includegraphics[height=1.45cm,width=1.45cm]{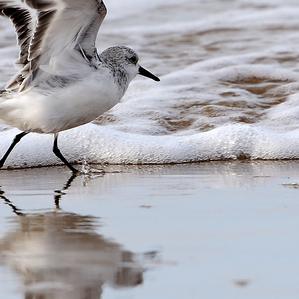}
		\end{minipage} 
	\text{\large{141}}
		\hfill  
		\begin{minipage}[t]{0.08\textwidth}
			\includegraphics[height=1.45cm,width=1.45cm]{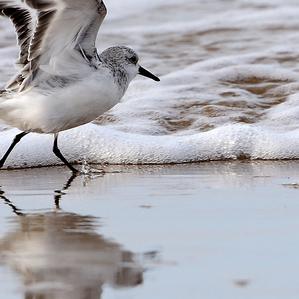}
		\end{minipage} 
		\hfill 
		\begin{minipage}[t]{0.08\textwidth}
			\includegraphics[height=1.45cm,width=1.45cm]{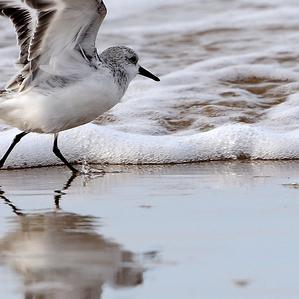}  
		\end{minipage} 
		\hfill  
		\begin{minipage}[t]{0.08\textwidth}
			\includegraphics[height=1.45cm,width=1.45cm]{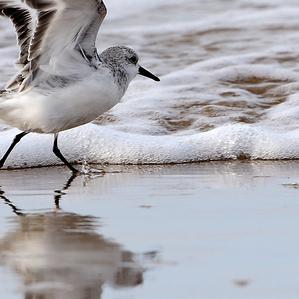}  
		\end{minipage} 
		\hfill  
		\begin{minipage}[t]{0.08\textwidth}
			\includegraphics[height=1.45cm,width=1.45cm]{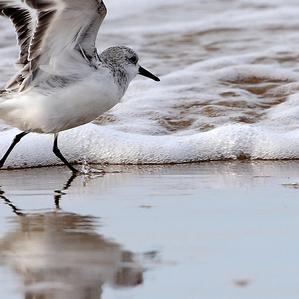}  
		\end{minipage} 
		\hfill  
		\begin{minipage}[t]{0.08\textwidth}
			\includegraphics[height=1.45cm,width=1.45cm]{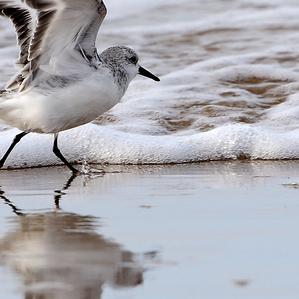}  
		\end{minipage} 
		\hfill  
		\begin{minipage}[t]{0.08\textwidth}
			\includegraphics[height=1.45cm,width=1.45cm]{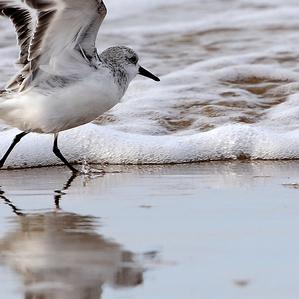}  
		\end{minipage} 
		\hfill  
		\begin{minipage}[t]{0.08\textwidth}
			\includegraphics[height=1.45cm,width=1.45cm]{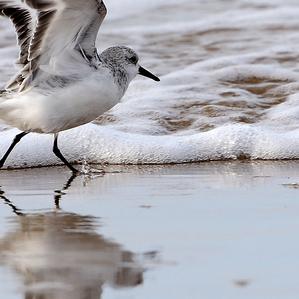}  
		\end{minipage} 
		\hfill  
		\begin{minipage}[t]{0.08\textwidth}
			\includegraphics[height=1.45cm,width=1.45cm]{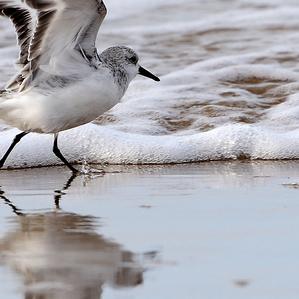}  
		\end{minipage} 
		\hfill  
		\begin{minipage}[t]{0.08\textwidth}
			\includegraphics[height=1.45cm,width=1.45cm]{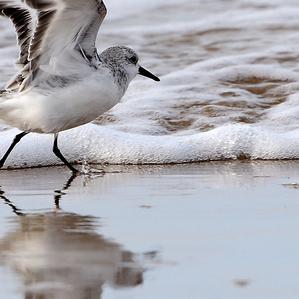}  
		\end{minipage} 
		\hfill  
		\begin{minipage}[t]{0.08\textwidth}
			\includegraphics[height=1.45cm,width=1.45cm]{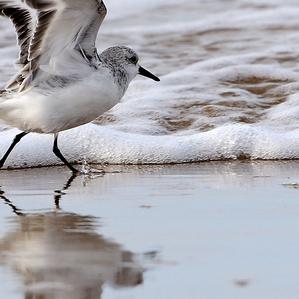} 
		\end{minipage} 
		\vfill
		\vspace{3.6pt}
		\begin{minipage}[t]{0.08\textwidth}
			\includegraphics[height=1.45cm,width=1.45cm]{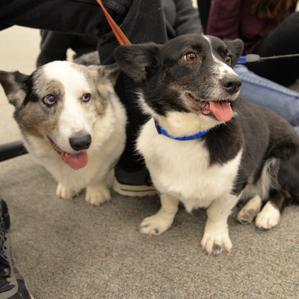}
		\end{minipage} 
		\text{\large{265}}
		\hfill  
		\begin{minipage}[t]{0.08\textwidth}
			\includegraphics[height=1.45cm,width=1.45cm]{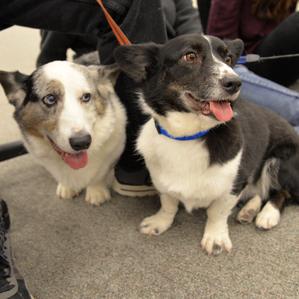}
		\end{minipage} 
		\hfill 
		\begin{minipage}[t]{0.08\textwidth}
			\includegraphics[height=1.45cm,width=1.45cm]{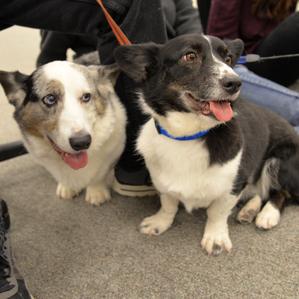}  
		\end{minipage} 
		\hfill  
		\begin{minipage}[t]{0.08\textwidth}
			\includegraphics[height=1.45cm,width=1.45cm]{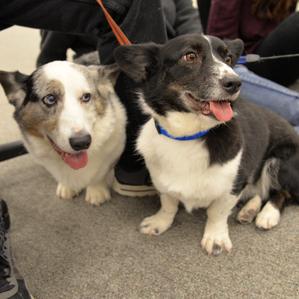}  
		\end{minipage} 
		\hfill  
		\begin{minipage}[t]{0.08\textwidth}
			\includegraphics[height=1.45cm,width=1.45cm]{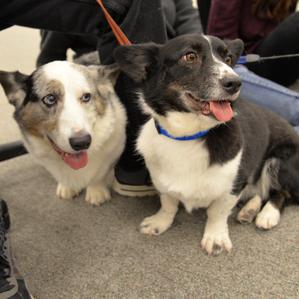}  
		\end{minipage} 
		\hfill  
		\begin{minipage}[t]{0.08\textwidth}
			\includegraphics[height=1.45cm,width=1.45cm]{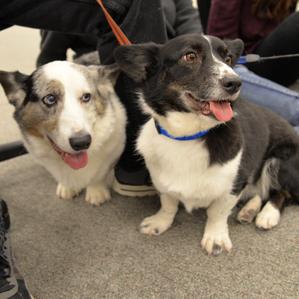}  
		\end{minipage} 
		\hfill  
		\begin{minipage}[t]{0.08\textwidth}
			\includegraphics[height=1.45cm,width=1.45cm]{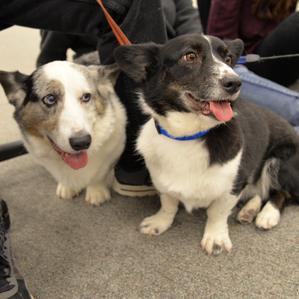}  
		\end{minipage} 
		\hfill  
		\begin{minipage}[t]{0.08\textwidth}
			\includegraphics[height=1.45cm,width=1.45cm]{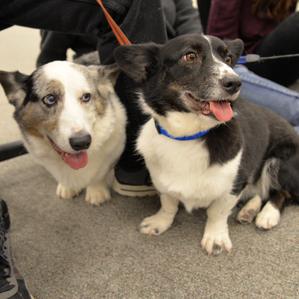}  
		\end{minipage} 
		\hfill  
		\begin{minipage}[t]{0.08\textwidth}
			\includegraphics[height=1.45cm,width=1.45cm]{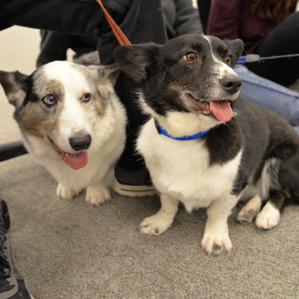}  
		\end{minipage} 
		\hfill  
		\begin{minipage}[t]{0.08\textwidth}
			\includegraphics[height=1.45cm,width=1.45cm]{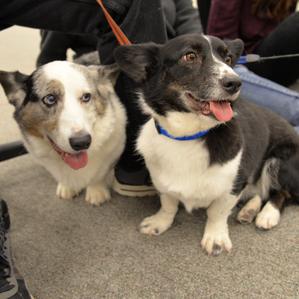}  
		\end{minipage} 
		\hfill  
		\begin{minipage}[t]{0.08\textwidth}
			\includegraphics[height=1.45cm,width=1.45cm]{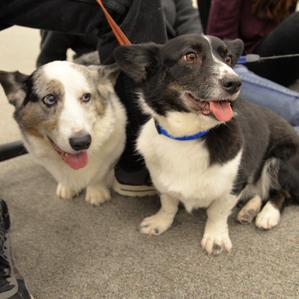} 
		\end{minipage} 
		\vfill
		\vspace{3.6pt}
		\begin{minipage}[t]{0.08\textwidth}
	\includegraphics[height=1.45cm,width=1.45cm]{imagenet_309_309.jpg}
\end{minipage} 
\text{\large{309}}
\hfill  
\begin{minipage}[t]{0.08\textwidth}
	\includegraphics[height=1.45cm,width=1.45cm]{imagenet_309_13.jpg}
\end{minipage} 
\hfill 
\begin{minipage}[t]{0.08\textwidth}
	\includegraphics[height=1.45cm,width=1.45cm]{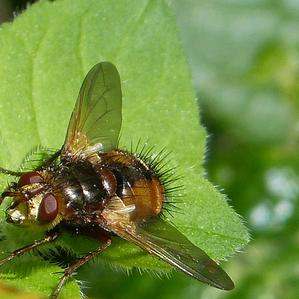}  
\end{minipage} 
\hfill  
\begin{minipage}[t]{0.08\textwidth}
	\includegraphics[height=1.45cm,width=1.45cm]{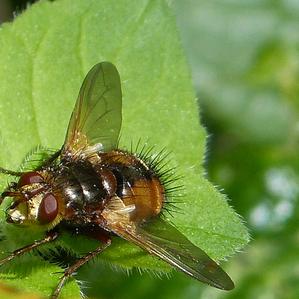}  
\end{minipage} 
\hfill  
\begin{minipage}[t]{0.08\textwidth}
	\includegraphics[height=1.45cm,width=1.45cm]{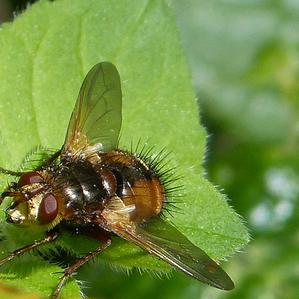}  
\end{minipage} 
\hfill  
\begin{minipage}[t]{0.08\textwidth}
	\includegraphics[height=1.45cm,width=1.45cm]{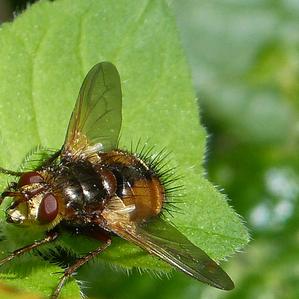}  
\end{minipage} 
\hfill  
\begin{minipage}[t]{0.08\textwidth}
	\includegraphics[height=1.45cm,width=1.45cm]{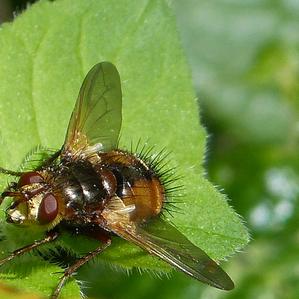}  
\end{minipage} 
\hfill  
\begin{minipage}[t]{0.08\textwidth}
	\includegraphics[height=1.45cm,width=1.45cm]{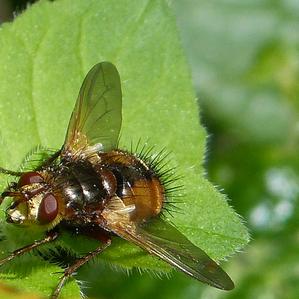}  
\end{minipage} 
\hfill  
\begin{minipage}[t]{0.08\textwidth}
	\includegraphics[height=1.45cm,width=1.45cm]{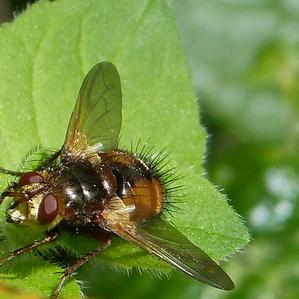}  
\end{minipage} 
\hfill  
\begin{minipage}[t]{0.08\textwidth}
	\includegraphics[height=1.45cm,width=1.45cm]{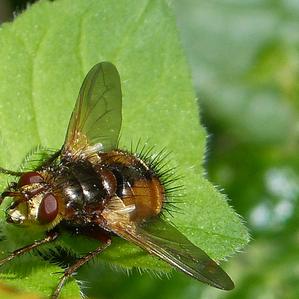}  
\end{minipage} 
\hfill  
\begin{minipage}[t]{0.08\textwidth}
	\includegraphics[height=1.45cm,width=1.45cm]{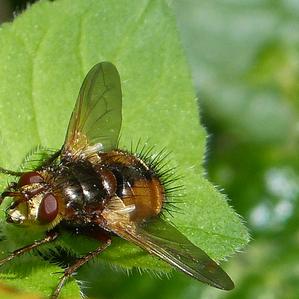} 
\end{minipage} 
\vfill
\vspace{3.6pt}
		\begin{minipage}[t]{0.08\textwidth}
	\includegraphics[height=1.45cm,width=1.45cm]{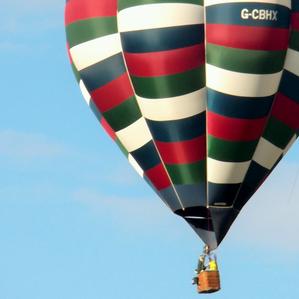}
\end{minipage} 
\text{\large{418}}
\hfill  
\begin{minipage}[t]{0.08\textwidth}
	\includegraphics[height=1.45cm,width=1.45cm]{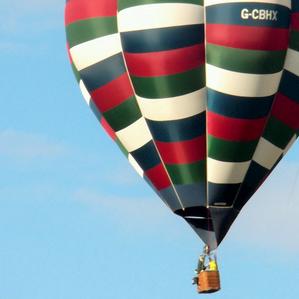}
\end{minipage} 
\hfill 
\begin{minipage}[t]{0.08\textwidth}
	\includegraphics[height=1.45cm,width=1.45cm]{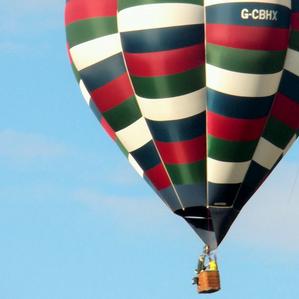}  
\end{minipage} 
\hfill  
\begin{minipage}[t]{0.08\textwidth}
	\includegraphics[height=1.45cm,width=1.45cm]{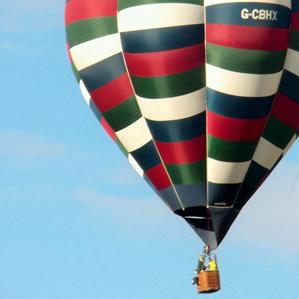}  
\end{minipage} 
\hfill  
\begin{minipage}[t]{0.08\textwidth}
	\includegraphics[height=1.45cm,width=1.45cm]{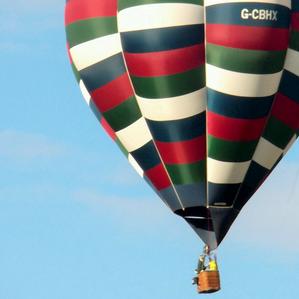}  
\end{minipage} 
\hfill  
\begin{minipage}[t]{0.08\textwidth}
	\includegraphics[height=1.45cm,width=1.45cm]{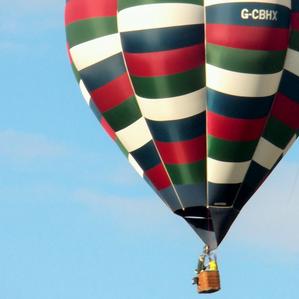}  
\end{minipage} 
\hfill  
\begin{minipage}[t]{0.08\textwidth}
	\includegraphics[height=1.45cm,width=1.45cm]{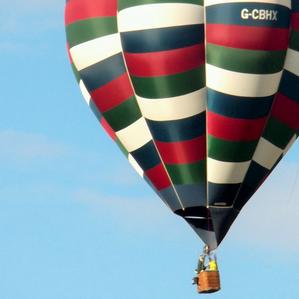}  
\end{minipage} 
\hfill  
\begin{minipage}[t]{0.08\textwidth}
	\includegraphics[height=1.45cm,width=1.45cm]{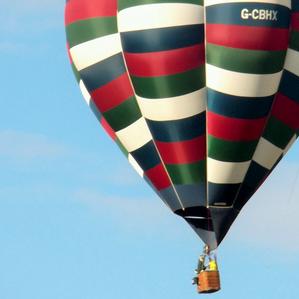}  
\end{minipage} 
\hfill  
\begin{minipage}[t]{0.08\textwidth}
	\includegraphics[height=1.45cm,width=1.45cm]{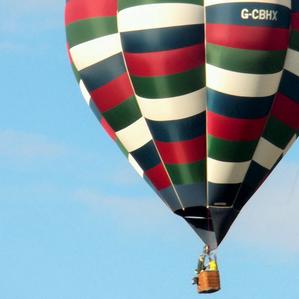}  
\end{minipage} 
\hfill  
\begin{minipage}[t]{0.08\textwidth}
	\includegraphics[height=1.45cm,width=1.45cm]{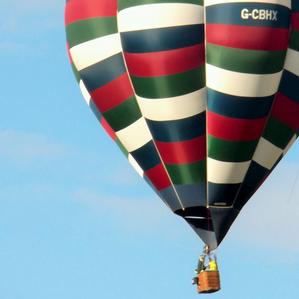}  
\end{minipage} 
\hfill  
\begin{minipage}[t]{0.08\textwidth}
	\includegraphics[height=1.45cm,width=1.45cm]{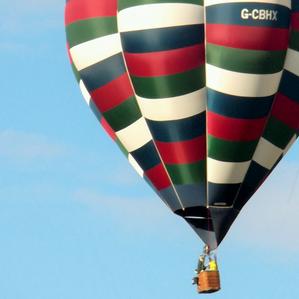} 
\end{minipage} 
\vfill
\vspace{3.6pt}
		\begin{minipage}[t]{0.08\textwidth}
	\includegraphics[height=1.45cm,width=1.45cm]{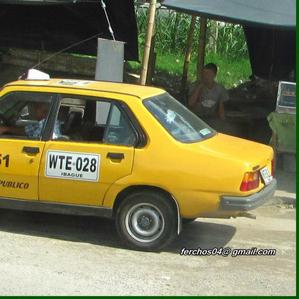}
\end{minipage} 
\text{\large{469}}
\hfill  
\begin{minipage}[t]{0.08\textwidth}
	\includegraphics[height=1.45cm,width=1.45cm]{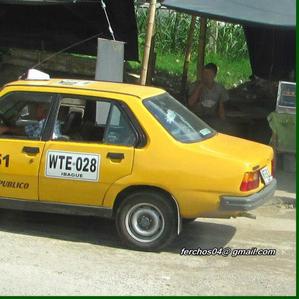}
\end{minipage} 
\hfill 
\begin{minipage}[t]{0.08\textwidth}
	\includegraphics[height=1.45cm,width=1.45cm]{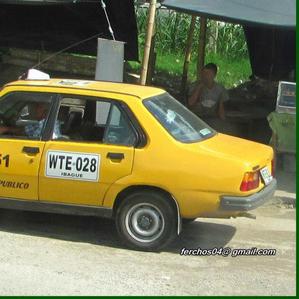}  
\end{minipage} 
\hfill  
\begin{minipage}[t]{0.08\textwidth}
	\includegraphics[height=1.45cm,width=1.45cm]{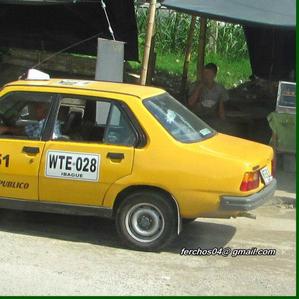}  
\end{minipage} 
\hfill  
\begin{minipage}[t]{0.08\textwidth}
	\includegraphics[height=1.45cm,width=1.45cm]{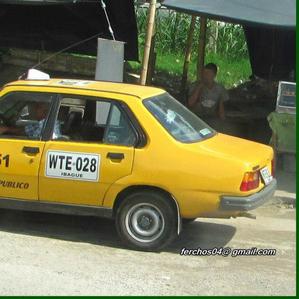}  
\end{minipage} 
\hfill  
\begin{minipage}[t]{0.08\textwidth}
	\includegraphics[height=1.45cm,width=1.45cm]{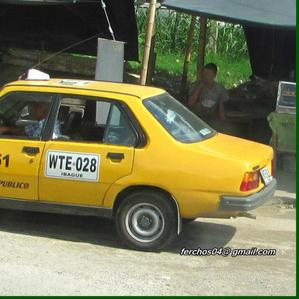}  
\end{minipage} 
\hfill  
\begin{minipage}[t]{0.08\textwidth}
	\includegraphics[height=1.45cm,width=1.45cm]{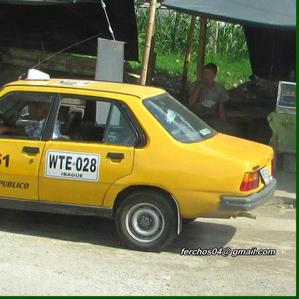}  
\end{minipage} 
\hfill  
\begin{minipage}[t]{0.08\textwidth}
	\includegraphics[height=1.45cm,width=1.45cm]{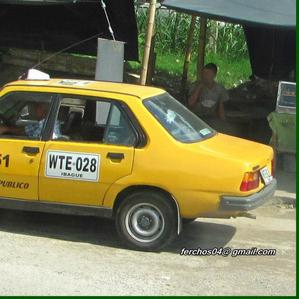}  
\end{minipage} 
\hfill  
\begin{minipage}[t]{0.08\textwidth}
	\includegraphics[height=1.45cm,width=1.45cm]{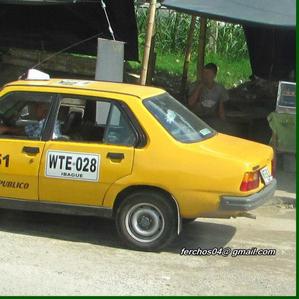}  
\end{minipage} 
\hfill  
\begin{minipage}[t]{0.08\textwidth}
	\includegraphics[height=1.45cm,width=1.45cm]{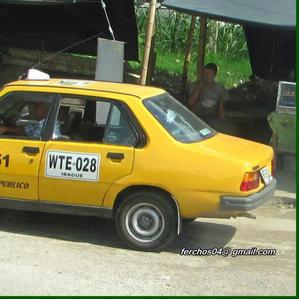}  
\end{minipage} 
\hfill  
\begin{minipage}[t]{0.08\textwidth}
	\includegraphics[height=1.45cm,width=1.45cm]{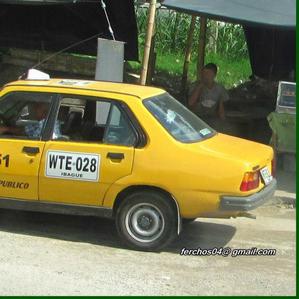} 
\end{minipage} 
\vfill
\vspace{3.6pt}
		\begin{minipage}[t]{0.08\textwidth}
	\includegraphics[height=1.45cm,width=1.45cm]{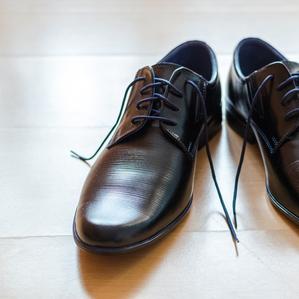}
\end{minipage} 
\text{\large{631}}
\hfill  
\begin{minipage}[t]{0.08\textwidth}
	\includegraphics[height=1.45cm,width=1.45cm]{imagenet_631_631.jpg}
\end{minipage} 
\hfill 
\begin{minipage}[t]{0.08\textwidth}
	\includegraphics[height=1.45cm,width=1.45cm]{imagenet_631_631.jpg}  
\end{minipage} 
\hfill  
\begin{minipage}[t]{0.08\textwidth}
	\includegraphics[height=1.45cm,width=1.45cm]{imagenet_631_631.jpg}  
\end{minipage} 
\hfill  
\begin{minipage}[t]{0.08\textwidth}
	\includegraphics[height=1.45cm,width=1.45cm]{imagenet_631_631.jpg}  
\end{minipage} 
\hfill  
\begin{minipage}[t]{0.08\textwidth}
	\includegraphics[height=1.45cm,width=1.45cm]{imagenet_631_631.jpg}  
\end{minipage} 
\hfill  
\begin{minipage}[t]{0.08\textwidth}
	\includegraphics[height=1.45cm,width=1.45cm]{imagenet_631_631.jpg}  
\end{minipage} 
\hfill  
\begin{minipage}[t]{0.08\textwidth}
	\includegraphics[height=1.45cm,width=1.45cm]{imagenet_631_631.jpg}  
\end{minipage} 
\hfill  
\begin{minipage}[t]{0.08\textwidth}
	\includegraphics[height=1.45cm,width=1.45cm]{imagenet_631_631.jpg}  
\end{minipage} 
\hfill  
\begin{minipage}[t]{0.08\textwidth}
	\includegraphics[height=1.45cm,width=1.45cm]{imagenet_631_631.jpg}  
\end{minipage} 
\hfill  
\begin{minipage}[t]{0.08\textwidth}
	\includegraphics[height=1.45cm,width=1.45cm]{imagenet_631_631.jpg} 
\end{minipage} 
\vfill
\vspace{3.6pt}
		\begin{minipage}[t]{0.08\textwidth}
	\includegraphics[height=1.45cm,width=1.45cm]{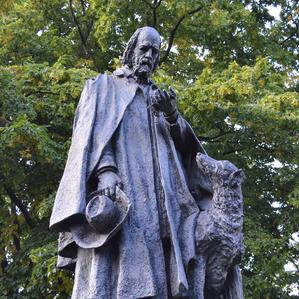}
\end{minipage} 
\text{\large{709}}
\hfill  
\begin{minipage}[t]{0.08\textwidth}
	\includegraphics[height=1.45cm,width=1.45cm]{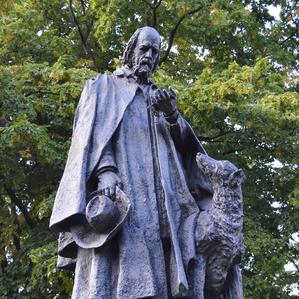}
\end{minipage} 
\hfill 
\begin{minipage}[t]{0.08\textwidth}
	\includegraphics[height=1.45cm,width=1.45cm]{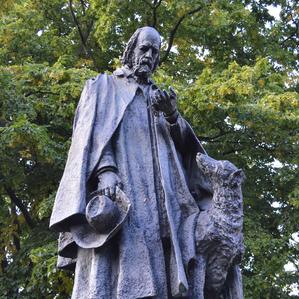}  
\end{minipage} 
\hfill  
\begin{minipage}[t]{0.08\textwidth}
	\includegraphics[height=1.45cm,width=1.45cm]{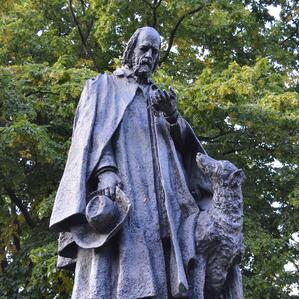}  
\end{minipage} 
\hfill  
\begin{minipage}[t]{0.08\textwidth}
	\includegraphics[height=1.45cm,width=1.45cm]{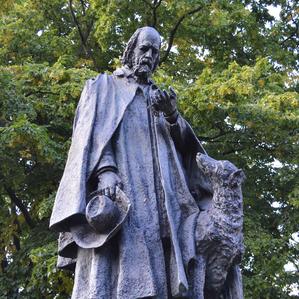}  
\end{minipage} 
\hfill  
\begin{minipage}[t]{0.08\textwidth}
	\includegraphics[height=1.45cm,width=1.45cm]{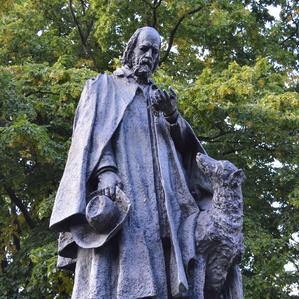}  
\end{minipage} 
\hfill  
\begin{minipage}[t]{0.08\textwidth}
	\includegraphics[height=1.45cm,width=1.45cm]{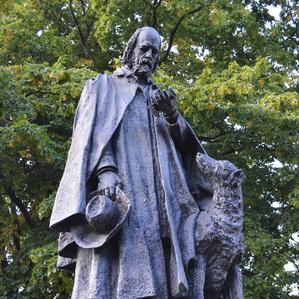}  
\end{minipage} 
\hfill  
\begin{minipage}[t]{0.08\textwidth}
	\includegraphics[height=1.45cm,width=1.45cm]{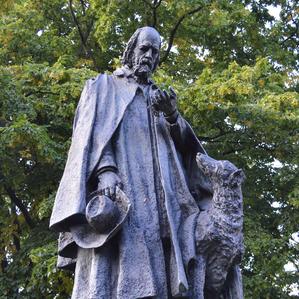}  
\end{minipage} 
\hfill  
\begin{minipage}[t]{0.08\textwidth}
	\includegraphics[height=1.45cm,width=1.45cm]{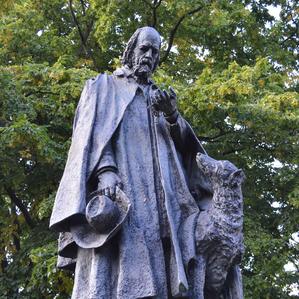}  
\end{minipage} 
\hfill  
\begin{minipage}[t]{0.08\textwidth}
	\includegraphics[height=1.45cm,width=1.45cm]{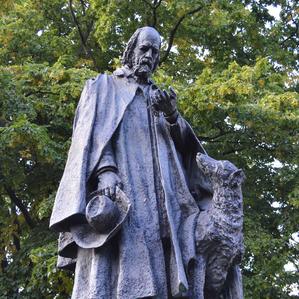}  
\end{minipage} 
\hfill  
\begin{minipage}[t]{0.08\textwidth}
	\includegraphics[height=1.45cm,width=1.45cm]{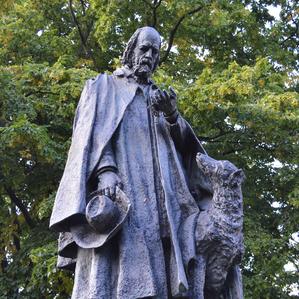} 
\end{minipage} 
\vfill
\vspace{3.6pt}
		\begin{minipage}[t]{0.08\textwidth}
	\includegraphics[height=1.45cm,width=1.45cm]{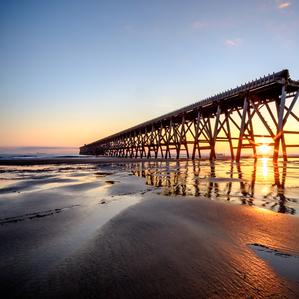}
\end{minipage} 
\text{\large{719}}
\hfill  
\begin{minipage}[t]{0.08\textwidth}
	\includegraphics[height=1.45cm,width=1.45cm]{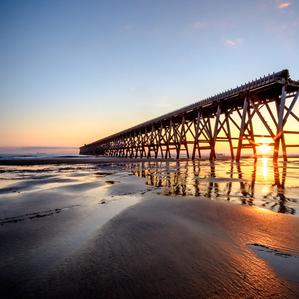}
\end{minipage} 
\hfill 
\begin{minipage}[t]{0.08\textwidth}
	\includegraphics[height=1.45cm,width=1.45cm]{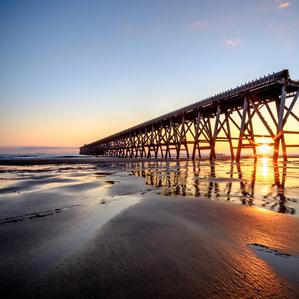}  
\end{minipage} 
\hfill  
\begin{minipage}[t]{0.08\textwidth}
	\includegraphics[height=1.45cm,width=1.45cm]{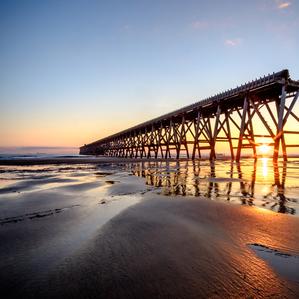}  
\end{minipage} 
\hfill  
\begin{minipage}[t]{0.08\textwidth}
	\includegraphics[height=1.45cm,width=1.45cm]{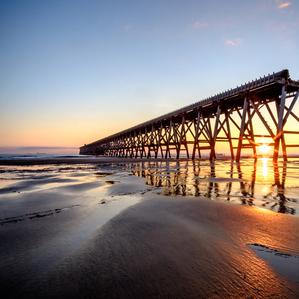}  
\end{minipage} 
\hfill  
\begin{minipage}[t]{0.08\textwidth}
	\includegraphics[height=1.45cm,width=1.45cm]{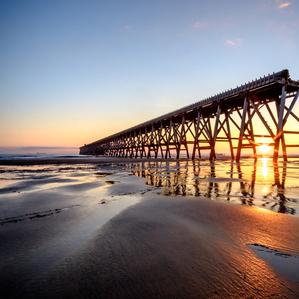}  
\end{minipage} 
\hfill  
\begin{minipage}[t]{0.08\textwidth}
	\includegraphics[height=1.45cm,width=1.45cm]{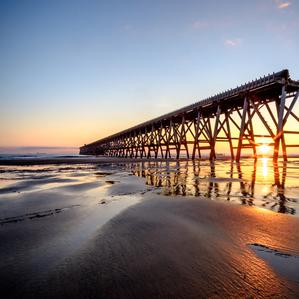}  
\end{minipage} 
\hfill  
\begin{minipage}[t]{0.08\textwidth}
	\includegraphics[height=1.45cm,width=1.45cm]{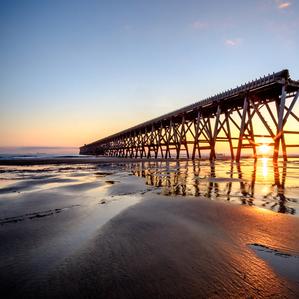}  
\end{minipage} 
\hfill  
\begin{minipage}[t]{0.08\textwidth}
	\includegraphics[height=1.45cm,width=1.45cm]{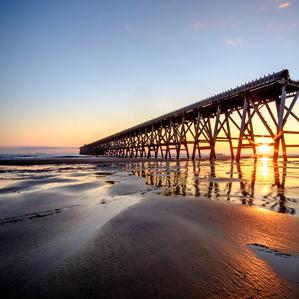}  
\end{minipage} 
\hfill  
\begin{minipage}[t]{0.08\textwidth}
	\includegraphics[height=1.45cm,width=1.45cm]{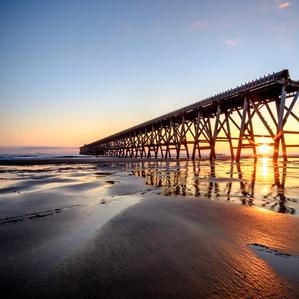}  
\end{minipage} 
\hfill  
\begin{minipage}[t]{0.08\textwidth}
	\includegraphics[height=1.45cm,width=1.45cm]{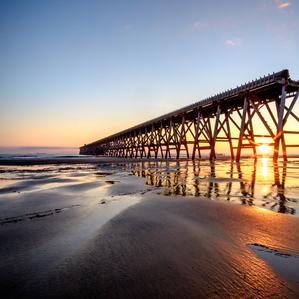} 
\end{minipage} 
\vfill
\vspace{3.6pt}
		\begin{minipage}[t]{0.08\textwidth}
	\includegraphics[height=1.45cm,width=1.45cm]{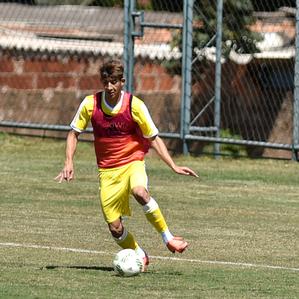}
\end{minipage} 
\text{\large{806}}
\hfill  
\begin{minipage}[t]{0.08\textwidth}
	\includegraphics[height=1.45cm,width=1.45cm]{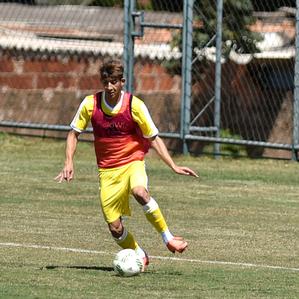}
\end{minipage} 
\hfill 
\begin{minipage}[t]{0.08\textwidth}
	\includegraphics[height=1.45cm,width=1.45cm]{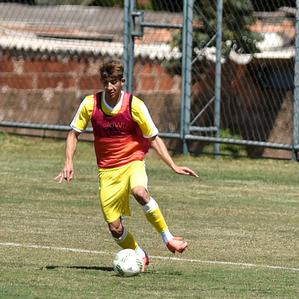}  
\end{minipage} 
\hfill  
\begin{minipage}[t]{0.08\textwidth}
	\includegraphics[height=1.45cm,width=1.45cm]{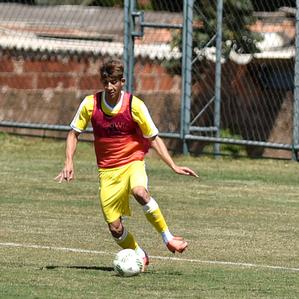}  
\end{minipage} 
\hfill  
\begin{minipage}[t]{0.08\textwidth}
	\includegraphics[height=1.45cm,width=1.45cm]{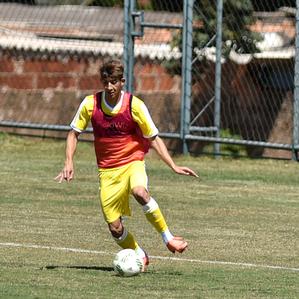}  
\end{minipage} 
\hfill  
\begin{minipage}[t]{0.08\textwidth}
	\includegraphics[height=1.45cm,width=1.45cm]{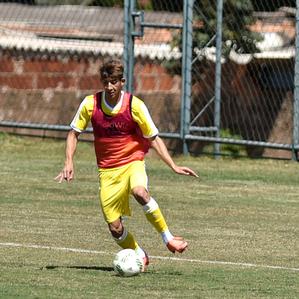}  
\end{minipage} 
\hfill  
\begin{minipage}[t]{0.08\textwidth}
	\includegraphics[height=1.45cm,width=1.45cm]{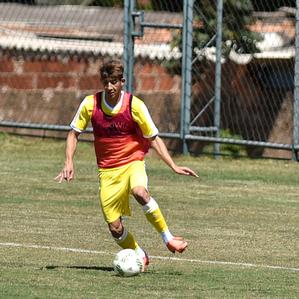}  
\end{minipage} 
\hfill  
\begin{minipage}[t]{0.08\textwidth}
	\includegraphics[height=1.45cm,width=1.45cm]{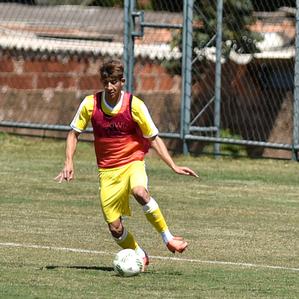}  
\end{minipage} 
\hfill  
\begin{minipage}[t]{0.08\textwidth}
	\includegraphics[height=1.45cm,width=1.45cm]{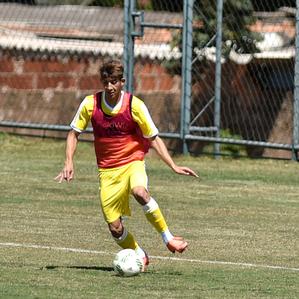}  
\end{minipage} 
\hfill  
\begin{minipage}[t]{0.08\textwidth}
	\includegraphics[height=1.45cm,width=1.45cm]{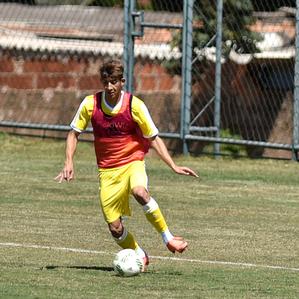}  
\end{minipage} 
\hfill  
\begin{minipage}[t]{0.08\textwidth}
	\includegraphics[height=1.45cm,width=1.45cm]{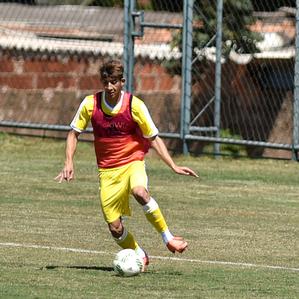} 
\end{minipage} 
\vfill
\vspace{3.6pt}
		\begin{minipage}[t]{0.08\textwidth}
	\includegraphics[height=1.45cm,width=1.45cm]{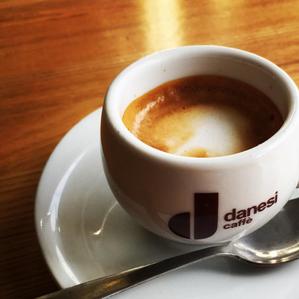}
\end{minipage} 
\text{\large{968}}
\hfill  
\begin{minipage}[t]{0.08\textwidth}
	\includegraphics[height=1.45cm,width=1.45cm]{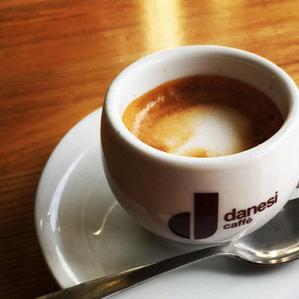}
\end{minipage} 
\hfill 
\begin{minipage}[t]{0.08\textwidth}
	\includegraphics[height=1.45cm,width=1.45cm]{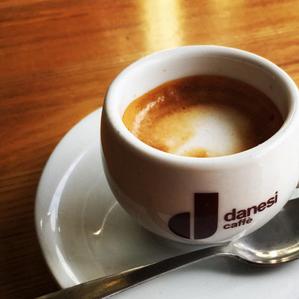}  
\end{minipage} 
\hfill  
\begin{minipage}[t]{0.08\textwidth}
	\includegraphics[height=1.45cm,width=1.45cm]{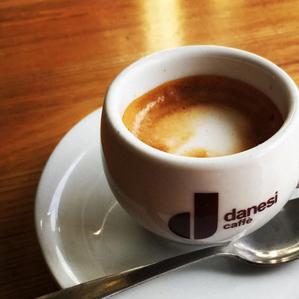}  
\end{minipage} 
\hfill  
\begin{minipage}[t]{0.08\textwidth}
	\includegraphics[height=1.45cm,width=1.45cm]{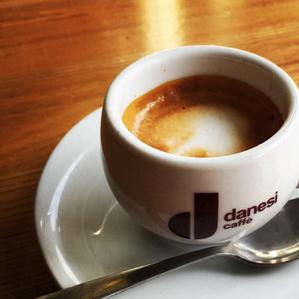}  
\end{minipage} 
\hfill  
\begin{minipage}[t]{0.08\textwidth}
	\includegraphics[height=1.45cm,width=1.45cm]{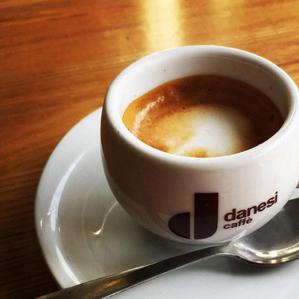}  
\end{minipage} 
\hfill  
\begin{minipage}[t]{0.08\textwidth}
	\includegraphics[height=1.45cm,width=1.45cm]{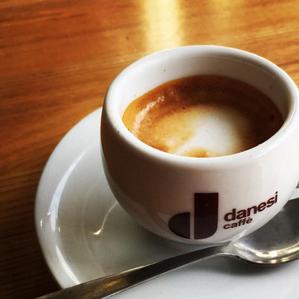}  
\end{minipage} 
\hfill  
\begin{minipage}[t]{0.08\textwidth}
	\includegraphics[height=1.45cm,width=1.45cm]{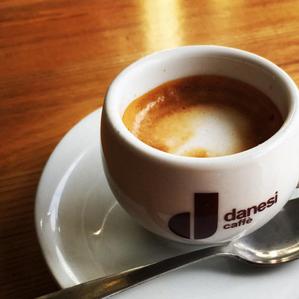}  
\end{minipage} 
\hfill  
\begin{minipage}[t]{0.08\textwidth}
	\includegraphics[height=1.45cm,width=1.45cm]{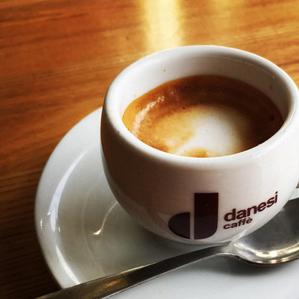}  
\end{minipage} 
\hfill  
\begin{minipage}[t]{0.08\textwidth}
	\includegraphics[height=1.45cm,width=1.45cm]{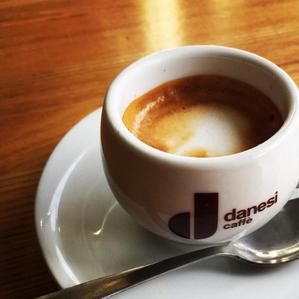}  
\end{minipage} 
\hfill  
\begin{minipage}[t]{0.08\textwidth}
	\includegraphics[height=1.45cm,width=1.45cm]{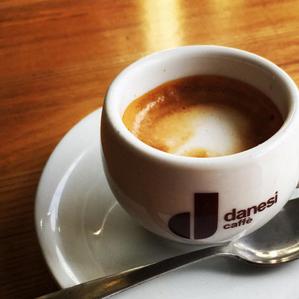} 
\end{minipage} 
\vfill
\vspace{3.6pt}

		\caption{\textbf{Targeted Adversarial Examples on ImagNet}. We randomly selected 10 images (in the first column), respectively belonging to 10 different classes, and the digits on the side are their original true labels. For each test image, we generate adversarial examples of 9 random selected classes as targets, as shown in each column.}
		\label{Fig:imagenet-show}
	\end{figure*}

The generated adversarial examples are shown in Fig.\ref{Fig:mnist-show} for MNIST, Fig.\ref{Fig:cifar-show} for CIFAR10, and Fig.\ref{Fig:imagenet-show} for ImageNet. Human eyes completely cannot tell the difference between the original images and adversarial examples on CIFAR10 and ImageNet. Here we compare our method with C\&W $\ell_{2}$ attack in the amount of the modification to images. The result is shown in Table \ref{table:Distortion-to-Images}. The max perturbation is the largest modification to all pixels of an image, and the value in the table is the mean max perturbation of all our test 9000 images. The total distortion is Euclidean distance between our adversarial examples and original images ($\ell_{2}$ norm), and we also record the mean value of 9000 images. From  Table \ref{table:Distortion-to-Images}, we can see that, when the success rates of the two attacks  are both 100\%, on MNIST and CIFAR10 dataset, our total distortion is larger than C\&W, but our max perturbation to an image is much smaller than that. The reason is that, in our optimization process, we have restrictions on the maximum modification, while in C\&W $\ell_{2}$, it is aimed at finding the minimum total distortion. Intuitively, several large perturbations to pixels in C\&W attack are spread to the entire image with our method. On ImageNet, we can find that in both aspects of max perturbation and total distortion, our $\epsilon$-neighborhood attack is better. The largest modification to each pixel of our attack is no more than 2, and the total distortion is also lower than C\&W. So our adversarial perturbation is less obvious and more imperceptible, as shown in Fig.\ref{Fig:cw-our}. In addition, we discover that, even if we preset the allowed maximum perturbation $\epsilon$, the final largest perturbation is usually smaller than that. It can be explained as follows: The $\epsilon$ is a searching space we limit for our optimization problem. A larger $\epsilon$ means we allow the algorithm to modify the original image in a larger range. But even so, it may not be necessary to modify till the preset maximum range $\epsilon$ in the actual optimization process. Therefore, the real maximum perturbation is usually smaller than $\epsilon$.

%
\renewcommand\arraystretch{1.5}
\begin{table}
	\caption{\textbf{Distortion} to Images of Our $\epsilon$-neighborhood Attack compared with C\&W Attack, when the success rates of two methods are both 100\%.}
	\label{table:Distortion-to-Images}
	\centering 
	\begin{tabular}{c|p{1cm}<{\centering}|p{1cm}<{\centering}|p{1cm}<{\centering}|p{1cm}<{\centering}}
		\hline
		&\multicolumn{2}{c|}{\textbf{Max Perturbation}}&
		\multicolumn{2}{c}{\textbf{Total Distortion}}\\
		\hline
		&Our $\epsilon$&C\&W&Our $\epsilon$&C\&W\\
		\hline
		MINST ($\epsilon$=52)&48.1&146.4&778.2&532.5\\
		\hline
		CIFAR10 ($\epsilon$=10)&5.6&11.6&194.5&107.6\\
		\hline
		ImageNet ($\epsilon$=2)&1.5&12.1&363.5&384.0\\
		\hline
	\end{tabular}
\end{table}

\subsubsection{Runtime}
Our $\epsilon$-neighborhood attack has a great advantage over the C\&W algorithm in terms of processing speed. As shown in Table \ref{table:Runtime}, our runtime is much shorter than C\&W attack on these three datasets.

\begin{table}
	\caption{\textbf{Runtime} Comparison of Generating Adversarial Examples on Three Datasets by Two Attacks.} 
	\label{table:Runtime}
	\centering 
	\begin{tabular}{c|c|c}
		\hline
		&Our $\epsilon$ (s /image)&C\&W (s /image)\\
		\hline
		MNIST ($\epsilon$=52)&0.53&2.11\\
		\hline
		CIFAR10 ($\epsilon$=10)&0.07&1.18\\
		\hline
		ImageNet ($\epsilon$=2)&4.27&79.80\\
		\hline
	\end{tabular}
\end{table}

\begin{figure}
	\centering  
	\includegraphics[height=2.7in, width=3.6in]{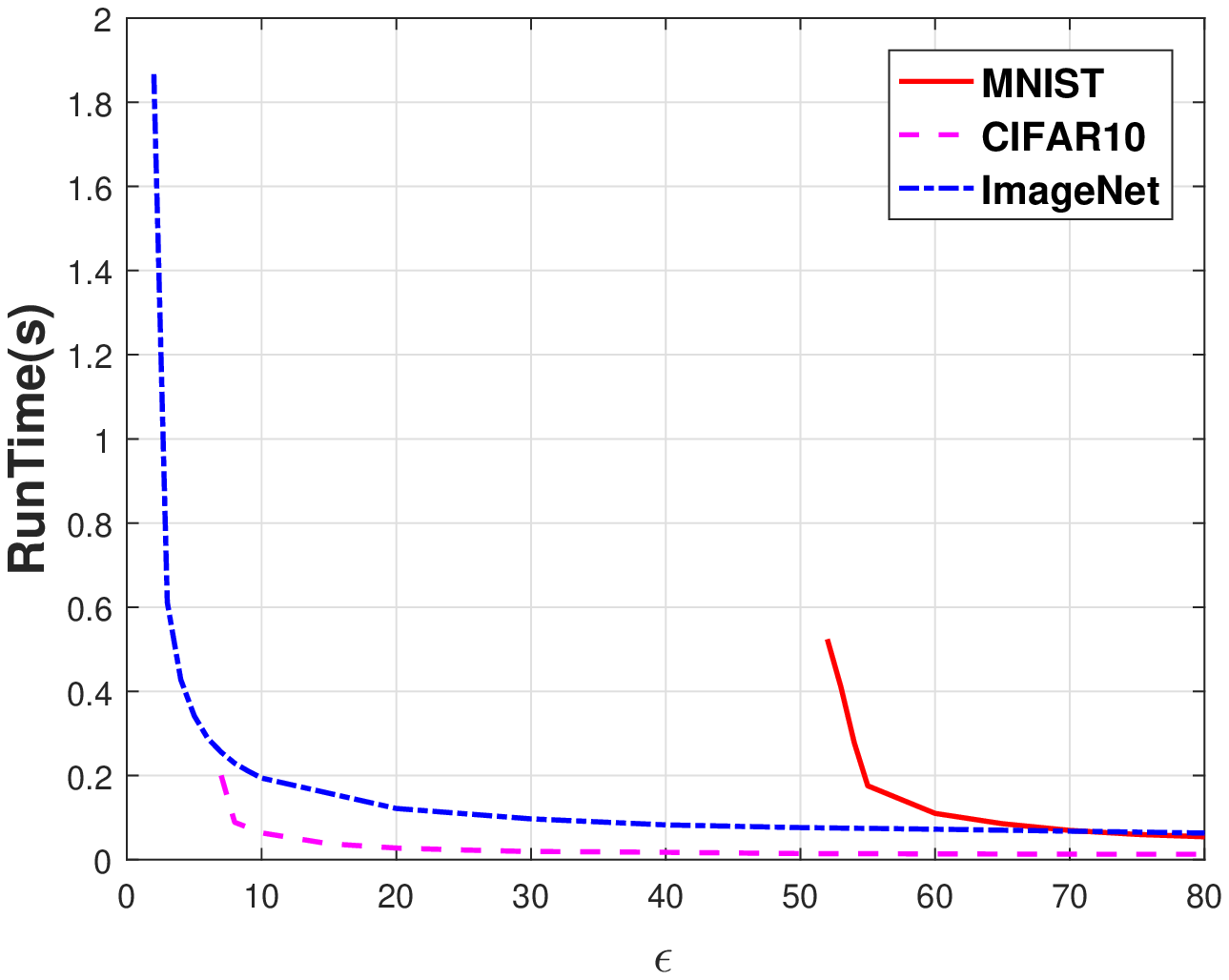}
	\caption{\textbf{Runtime with Different $\mathbf{\epsilon}$.} It descends rapidly when $\mathbf{\epsilon}$ increases on the three datasets. The three curves all begin with the $\epsilon$ which makes the attack success rate reaches 100\%.}
	\label{Fig:epsilon-runtime}
\end{figure}

If we can tolerate a larger $\epsilon$, the processing speed will be faster. The more perturbation we allow to add, the more easily the optimal solution will be found. Thus the runtime will be shorter. It is a tradeoff between $\epsilon$ and runtime, as shown in Fig.\ref{Fig:epsilon-runtime}. The three curves all start from the $\epsilon$ which makes the attack success rate reach 100\%. We can find that, when the success rate has just reached 100\% with a certain $\epsilon$, the runtime may be a little long. But at this time, we only need to slightly increase $\epsilon$, the runtime will decline sharply, and the visual quality will not deteriorate. If attackers hope to generate adversarial examples faster, they can set a slightly larger $\epsilon$. In a word, with the method of our $\epsilon$-neighborhood attack, the attackers can choose $\epsilon$ as needed, while the process time of all the previous attacks cannot be controlled by attackers.  

\subsection{Performance of Attacks on Black-box Distilled Model}
\begin{figure}
	\centering  
	\includegraphics[height=2.7in, width=3.6in]{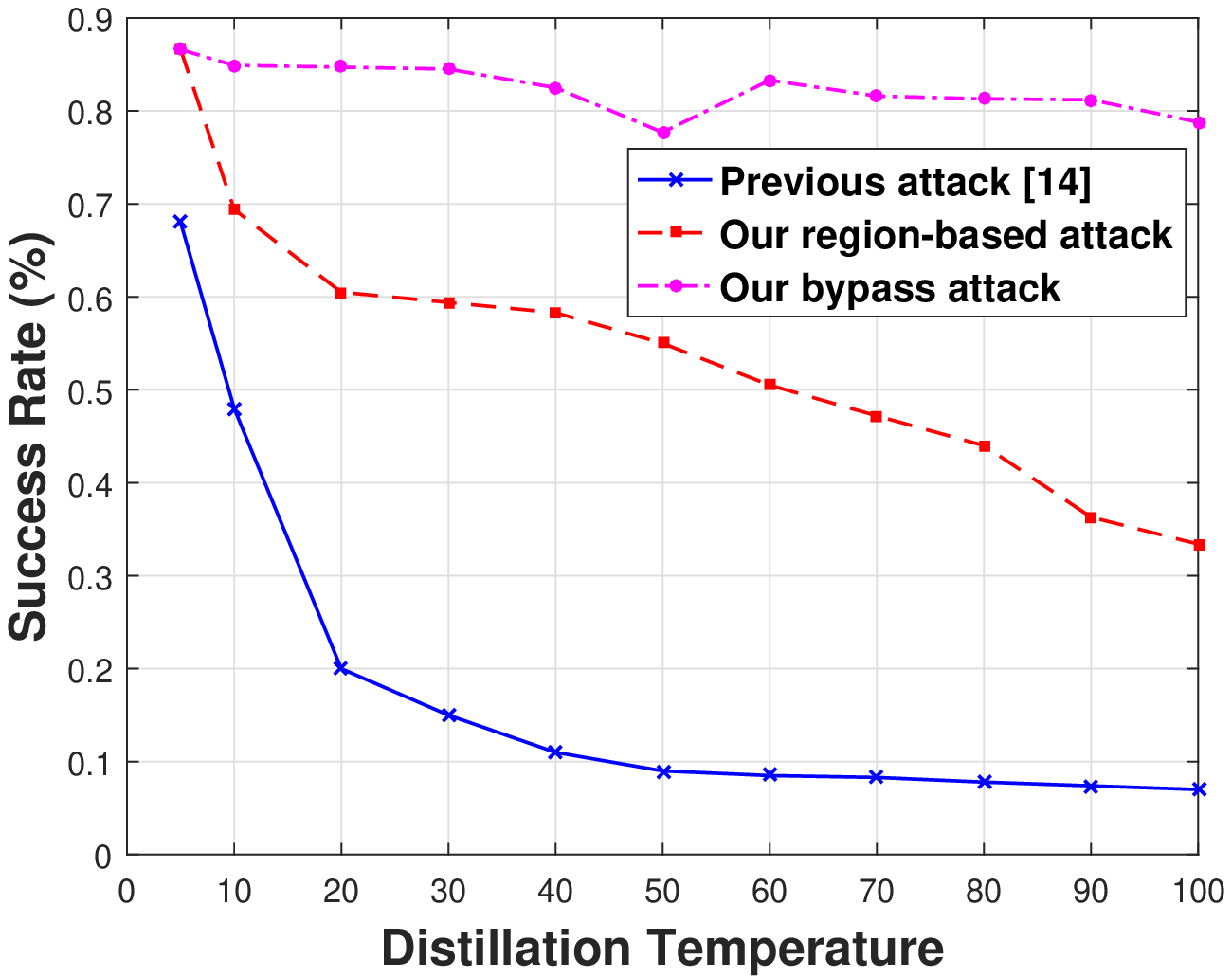}
	\caption{\textbf{The Effectiveness of Our Black-box Attack}. The previous attack is decribed in \cite{papernot2016limitations}. We generate adversarial examples on the distilled network with $T=\{5,10,20,...,100\}$ by our region-based attack. And we implement our bypass attack by constructing adversarial examples on models with $T=5$ and using them to attack distilled networks with other various $T$.}
	\label{Fig:direct-distill}
\end{figure}

Aimed at black-box distilled models, we propose two attacks. First we evaluate the performance of our region-based attack. We generate adversarial examples on black-box networks with $T=\{5,10,20,...,100\}$, the success rates are shown in Fig.\ref{Fig:direct-distill}. Compared with the previous attack described in \cite{papernot2016limitations}, the performance of our region-based attack has greatly improved. It shows that the idea of transforming the point-based optimization into region-based optimization is effective. Although distillation operation makes the output probability so ``hard" that the optimization becomes very difficult, our region-based attack soften it by constantly adding noise during optimization. We make it possible to produce adversarial examples on ``hard'' probability. 

However, when $T$ is very large ($T=100$), the success rate is still unsatisfactory (about $35\%$). It indicates that distilled networks with great $T$ are too robust for the region-based method to attack. Under these circumstances, we utilize the transferability of adversarial examples on models with different distillation temperatures to achieve our bypass attack. This circuitous mechanism is more effective than the direct method. We craft adversarial examples on a distilled model with low temperature, and use them to attack the models with high temperatures. Our experiments show that the attack success rate at other high temperatures can almost achieve the same level as the low temperature itself, as shown in Fig.\ref{Fig:direct-distill}. There is a strong transferability among models with different $T$. When adversarial examples generated for the network with a certain temperature are tested at other temperatures, the success rates are usually similar, whether they are high or low, shown in Table \ref{table:Bypass-attack}, the values in a row are very close.

\begin{table*}
	\caption{\textbf{Bypass Attack between Different Distillation Temperature.} We directly generate adversarial examples on distilled models with $T=\{5,10,20,70,100\}$, and we test the generated adversarial examples on models with a large range of temperatures $T=\{5,10,20,...,100\}$.} 
	\label{table:Bypass-attack}
	\centering 
	\begin{tabular}{c|ccccccccccc}
		\hline
		&$T=5$&$T=10$&$T=20$&$T=30$&$T=40$&$T=50$&$T=60$&$T=70$&$T=80$&$T=90$&$T=100$\\
		\hline
		\rowcolor{gray!60}$T=5$&0.866&0.844&0.844&0.844&0.822&0.777&0.833&0.811&0.811&0.811&0.788\\
		\rowcolor{gray!40}$T=10$&0.644&0.644&0.666&0.6&0.6&0.555&0.6&0.533&0.577&0.622&0.555\\
		\rowcolor{gray!20}$T=20$&0.466&0.488&0.555&0.466&0.444&0.444&0.488&0.444&0.466&0.533&0.466\\
		\rowcolor{gray!5}$T=70$&0.377&0.4&0.377&0.377&0.377&0.288&0.4&0.422&0.333&0.422&0.355\\
		\rowcolor{gray!5}$T=100$&0.266&0.311&0.244&0.266&0.244&0.266&0.0.244&0.244&0.244&0.266&0.266\\
		\hline
	\end{tabular}
\end{table*}

In our two attacks on black-box distilled models, the noise intensity is a quite important parameter. Since we transform the point-based problem into a region-based problem by adding Gaussian noise in each iteration of optimization, the noise intensity can be just thought of the radius of the region. Here we use the standard deviation $\sigma$ of added Gaussian noise to measure its intensity. Different noise in our attack will cause a big difference in the final success rate. 

\begin{figure}[t]
	\centering  
	\includegraphics[height=2.7in, width=3.6in]{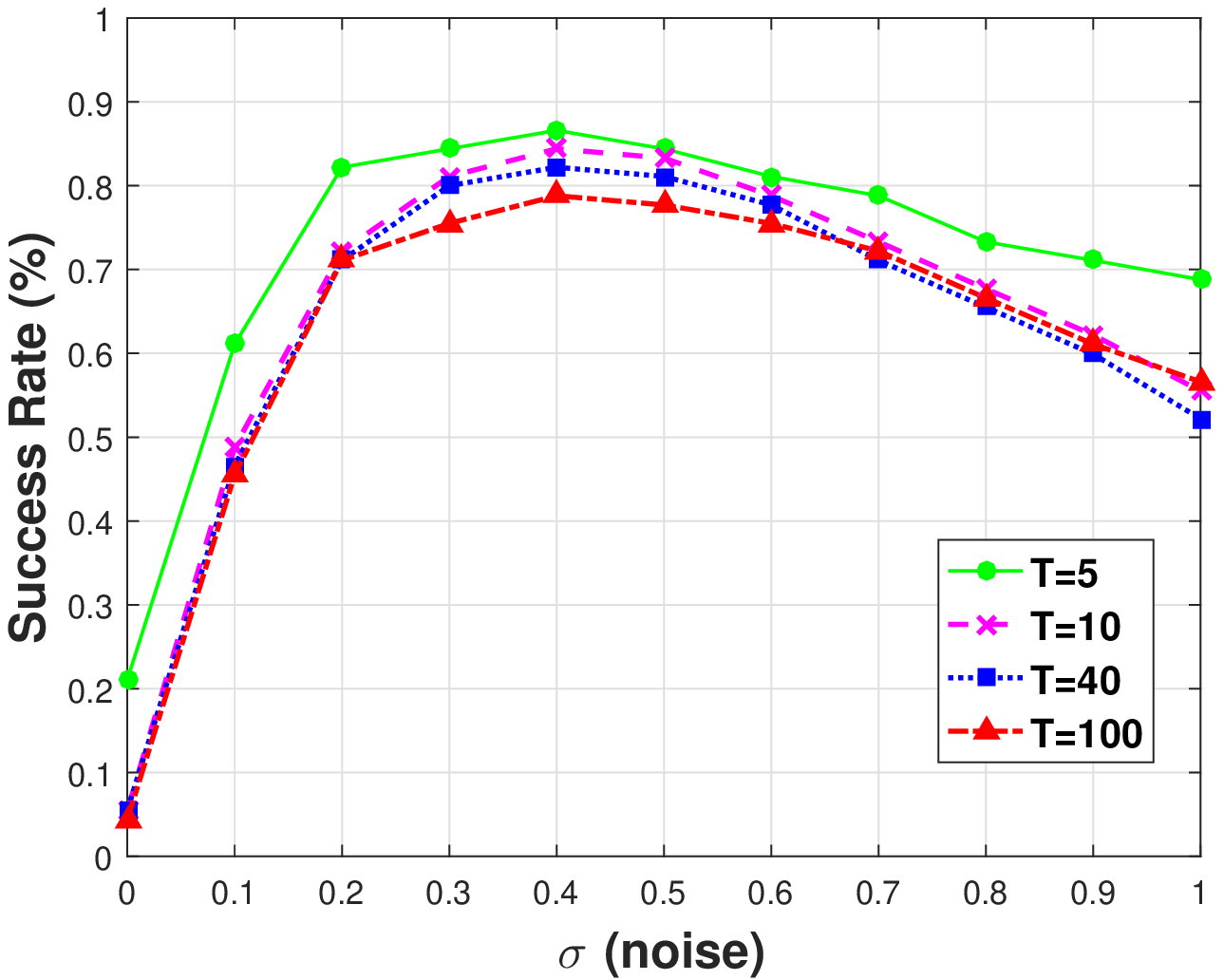}
	\caption{\textbf{Success Rate under Noise with Different Intensity in Region-based Attack.} When we use Gaussian noise with different standard deviation $\sigma$ in the optimization process, the final effectiveness is also different. We directly attack the model with distillation temperature $T=5$, and test the adversarial examples on models with $T=10,40,100$.}
	\label{Fig:bypass-noise-sr}
\end{figure}

As shown in Fig.\ref{Fig:bypass-noise-sr}, we add Gaussian noise with standard deviation in range of [0, 1.0]. We attack directly on the distilled model with $T=5$, and then use the generated adversarial examples to test on models with other $T$. For the distillation temperature $T=5$, the best noise is $\sigma=0.4$. The role of the noise is to make the corresponding point of the image deviate from its original position in the hyperspace. If we do not add noise or the  noise intensity is too low, the point will not move, so the optimization is difficult to be carried on. We notice that when $\sigma=0$, that is to say, we do not add any noise and directly attack black-box models in the same way with white-box $\epsilon$-neighborhood attack, only replace the logits with classification probability, the success rate is about $20\%$. But if we add a little noise ($\sigma=0.4$), the success rate will increase to over $85\%$. This indicates the effectiveness of our region-based idea. When $\sigma$ is too large, too much noise will be added in each optimization iteration. This makes it quite difficult to search an overall optimal solution, and the loss cannot decrease to below 0, so the attack will not be successful.

Therefore in our black-box attack, noise intensity $\sigma$ is a critical parameter that determines whether it will be successful. And for different temperature $T$, the optimal $\sigma$ may be different, we can adopt a certain search algorithm, \textit{e.g.} binary search, to find a suitable one. In addition, the effect of noise on our black-box bypass attack has the same trend as that on our direct region-based attack, as shown in Fig.
\ref{Fig:bypass-noise-sr}. If a certain $\sigma$ causes a good performance of direct attack, it will also work well at other temperatures by our bypass attack.

Better still, to a certain extent, our black-box attacks can be resistant to noise. Defensers may try to add noise onto the generated adversarial examples for damaging them and making them recover to the original correct label. If the noise intensity they add is no more than that in the process of our black-box attack, the classification results will still remain the targeted labels. The success rate of adding noise on adversarial examples can be as high as those without noise. From a different perspective, given an image, we do not just find one adversarial example, but find a lot. In the hyperspace, points in a region centered on the point corresponding to the generated adversarial example are all adversarial examples as well. This kind of robust anti-noise adversarial example may be a powerful opponent for defenders.

\subsection{Discussion on Our Black-box Attack}
The analysis above shows that we can combine our region-based attack and bypass attack in the black-box setting. When the distillation temperature $T$ of model is not high, the region-based attack can succeed directly; otherwise, the bypass attack can be used by generating adversarial examples on a small-$T$ model, without knowing any result of inner layers in the network. Actually, our bypass attack is not, strictly speaking, under black-box setting. Because if $T$ of the model is very large, we cannot succeed in attacking directly. So we need to produce adversarial examples with the assistance of another model with a small $T$. But we usually do not have the auxiliary model. The black-box here just refers that we only need the final output probabilities instead of the inner layers' results. 

Despite this, our bypass attack can still be applied in some specific scenarios. For example, many companies provide machine-learning-as-a-service (MLaaS) classifiers, from which users can get classification results when they input an image, or users can develop other applications via their prediction APIs. MLaaS is a black-box model and the model itself might be commercially valuable to model owners, so it just provides query operation. As we mentioned above, distillation can improve the robustness and security of a model and make it hard to be attacked. When the distillation temperature goes up, the security will be enhanced but model's accuracy will drop. So a company might deploy multiple distilled models with different temperatures on several clouds to meet various user requirements for both utility and security. Under these circumstances, attackers can easily use bypass attack to generate adversarial examples on low-temperature distilled model (utility first) and attack the high-temperature one (security first), even though which has such a high level of security that it is almost impossible to attack directly.

On another scenario, for a better tradeoff between utility and security, MLaaS providers need to train an optimal distilled model from an original undistilled model by constantly increasing the temperature. Users might have access to the historical models, that is, the trained models with not high temperatures. Then the final optimal model can be attacked by our bypass method. Or MLaaS providers might upgrade their old model with low temperature into a high-temperature one in order to enhance security. In this case, the adversarial examples generated on low-temperature model can be saved and used to attack the newly enhanced models effectively. In short, we conclude that, in the black-box  MLaaS, with the method of bypass attack, once the original model or the distilled models with low temperatures was released sometime in the past, its corresponding enhanced model will be not secure anymore, no matter what a high distillation temperature is used to upgrade the model.

\section{Conclusion}

The existence of adversarial examples proves that DNNs are not robust to the input. And this vulnerability hinders the deployment of DNN systems.
Defensively distilled DNN is currently the most effective defense to restrain the generation of adversarial examples. Although the state-of-the-art C\&W attack can defeat it with 100\% probability in the white-box setting, it still does not work in the more general black-box setting. 

In this paper, we first present the $\epsilon$-neighborhood attack. This method enables us to control the maximum perturbation of each pixel in an image, thus ensuring a high visual quality of adversarial examples. Also, we show that our method can find the targeted perturbations very fast.
On this basis, we further propose the region-based and the bypass attack, which show that the distillation does not provide enough security assurance to DNNs. An attacker can still generate adversarial examples with a high success rate, even though he only knows the final output probabilities of the target DNN.





\balance
\bibliographystyle{IEEEtran}
\bibliography{bare-conf-reference}

\begin{thebibliography}{10}
\providecommand{\url}[1]{#1}
\csname url@samestyle\endcsname
\providecommand{\newblock}{\relax}
\providecommand{\bibinfo}[2]{#2}
\providecommand{\BIBentrySTDinterwordspacing}{\spaceskip=0pt\relax}
\providecommand{\BIBentryALTinterwordstretchfactor}{4}
\providecommand{\BIBentryALTinterwordspacing}{\spaceskip=\fontdimen2\font plus
\BIBentryALTinterwordstretchfactor\fontdimen3\font minus
  \fontdimen4\font\relax}
\providecommand{\BIBforeignlanguage}[2]{{%
\expandafter\ifx\csname l@#1\endcsname\relax
\typeout{** WARNING: IEEEtran.bst: No hyphenation pattern has been}%
\typeout{** loaded for the language `#1'. Using the pattern for}%
\typeout{** the default language instead.}%
\else
\language=\csname l@#1\endcsname
\fi
#2}}
\providecommand{\BIBdecl}{\relax}
\BIBdecl

\bibitem{krizhevsky2012imagenet}
A.~Krizhevsky, I.~Sutskever, and G.~E. Hinton, ``Imagenet classification with
  deep convolutional neural networks,'' in \emph{Advances in neural information
  processing systems}, 2012, pp. 1097--1105.

\bibitem{szegedy2015going}
C.~Szegedy, W.~Liu, Y.~Jia, P.~Sermanet, S.~Reed, D.~Anguelov, D.~Erhan,
  V.~Vanhoucke, and A.~Rabinovich, ``Going deeper with convolutions,'' in
  \emph{Proceedings of the IEEE conference on computer vision and pattern
  recognition}, 2015, pp. 1--9.

\bibitem{simonyan2014very}
K.~Simonyan and A.~Zisserman, ``Very deep convolutional networks for
  large-scale image recognition,'' \emph{arXiv preprint arXiv:1409.1556}, 2014.

\bibitem{hinton2012deep}
G.~Hinton, L.~Deng, D.~Yu, G.~E. Dahl, A.-r. Mohamed, N.~Jaitly, A.~Senior,
  V.~Vanhoucke, P.~Nguyen, T.~N. Sainath \emph{et~al.}, ``Deep neural networks
  for acoustic modeling in speech recognition: The shared views of four
  research groups,'' \emph{IEEE Signal Processing Magazine}, vol.~29, no.~6,
  pp. 82--97, 2012.

\bibitem{mikolov2010recurrent}
T.~Mikolov, M.~Karafi{\'a}t, L.~Burget, J.~Cernock{\`y}, and S.~Khudanpur,
  ``Recurrent neural network based language model.'' in \emph{Interspeech},
  vol.~2, 2010, p.~3.

\bibitem{kriegeskorte2015deep}
N.~Kriegeskorte, ``Deep neural networks: a new framework for modeling
  biological vision and brain information processing,'' \emph{Annual Review of
  Vision Science}, vol.~1, pp. 417--446, 2015.

\bibitem{liang2014deep}
Z.~Liang, G.~Zhang, J.~X. Huang, and Q.~V. Hu, ``Deep learning for healthcare
  decision making with emrs,'' in \emph{Bioinformatics and Biomedicine (BIBM),
  2014 IEEE International Conference on}.\hskip 1em plus 0.5em minus
  0.4em\relax IEEE, 2014, pp. 556--559.

\bibitem{szegedy2013intriguing}
C.~Szegedy, W.~Zaremba, I.~Sutskever, J.~Bruna, D.~Erhan, I.~Goodfellow, and
  R.~Fergus, ``Intriguing properties of neural networks,'' \emph{arXiv preprint
  arXiv:1312.6199}, 2013.

\bibitem{bojarski2016end}
M.~Bojarski, D.~Del~Testa, D.~Dworakowski, B.~Firner, B.~Flepp, P.~Goyal, L.~D.
  Jackel, M.~Monfort, U.~Muller, J.~Zhang \emph{et~al.}, ``End to end learning
  for self-driving cars,'' \emph{arXiv preprint arXiv:1604.07316}, 2016.

\bibitem{parkhi2015deep}
O.~M. Parkhi, A.~Vedaldi, A.~Zisserman \emph{et~al.}, ``Deep face
  recognition.'' in \emph{BMVC}, vol.~1, no.~3, 2015, p.~6.

\bibitem{dahl2013large}
G.~E. Dahl, J.~W. Stokes, L.~Deng, and D.~Yu, ``Large-scale malware
  classification using random projections and neural networks,'' in
  \emph{Acoustics, Speech and Signal Processing (ICASSP), 2013 IEEE
  International Conference on}.\hskip 1em plus 0.5em minus 0.4em\relax IEEE,
  2013, pp. 3422--3426.

\bibitem{goodfellow2014explaining}
I.~J. Goodfellow, J.~Shlens, and C.~Szegedy, ``Explaining and harnessing
  adversarial examples,'' \emph{arXiv preprint arXiv:1412.6572}, 2014.

\bibitem{rozsa2016adversarial}
A.~Rozsa, E.~M. Rudd, and T.~E. Boult, ``Adversarial diversity and hard
  positive generation,'' in \emph{Proceedings of the IEEE Conference on
  Computer Vision and Pattern Recognition Workshops}, 2016, pp. 25--32.

\bibitem{papernot2016limitations}
N.~Papernot, P.~McDaniel, S.~Jha, M.~Fredrikson, Z.~B. Celik, and A.~Swami,
  ``The limitations of deep learning in adversarial settings,'' in
  \emph{Security and Privacy (EuroS\&P), 2016 IEEE European Symposium
  on}.\hskip 1em plus 0.5em minus 0.4em\relax IEEE, 2016, pp. 372--387.

\bibitem{moosavi2016deepfool}
S.-M. Moosavi-Dezfooli, A.~Fawzi, and P.~Frossard, ``Deepfool: a simple and
  accurate method to fool deep neural networks,'' in \emph{Proceedings of the
  IEEE Conference on Computer Vision and Pattern Recognition}, 2016, pp.
  2574--2582.

\bibitem{carlini2017towards}
N.~Carlini and D.~Wagner, ``Towards evaluating the robustness of neural
  networks,'' in \emph{Security and Privacy (SP), 2017 IEEE Symposium
  on}.\hskip 1em plus 0.5em minus 0.4em\relax IEEE, 2017, pp. 39--57.

\bibitem{lu2017no}
J.~Lu, H.~Sibai, E.~Fabry, and D.~Forsyth, ``No need to worry about adversarial
  examples in object detection in autonomous vehicles,'' \emph{arXiv preprint
  arXiv:1707.03501}, 2017.

\bibitem{das2017keeping}
N.~Das, M.~Shanbhogue, S.-T. Chen, F.~Hohman, L.~Chen, M.~E. Kounavis, and
  D.~H. Chau, ``Keeping the bad guys out: Protecting and vaccinating deep
  learning with jpeg compression,'' \emph{arXiv preprint arXiv:1705.02900},
  2017.

\bibitem{cao2017mitigating}
X.~Cao and N.~Z. Gong, ``Mitigating evasion attacks to deep neural networks via
  region-based classification,'' \emph{arXiv preprint arXiv:1709.05583}, 2017.

\bibitem{gu2014towards}
S.~Gu and L.~Rigazio, ``Towards deep neural network architectures robust to
  adversarial examples,'' \emph{arXiv preprint arXiv:1412.5068}, 2014.

\bibitem{papernot2016distillation}
N.~Papernot, P.~McDaniel, X.~Wu, S.~Jha, and A.~Swami, ``Distillation as a
  defense to adversarial perturbations against deep neural networks,'' in
  \emph{Security and Privacy (SP), 2016 IEEE Symposium on}.\hskip 1em plus
  0.5em minus 0.4em\relax IEEE, 2016, pp. 582--597.

\bibitem{kurakin2016adversarial}
A.~Kurakin, I.~Goodfellow, and S.~Bengio, ``Adversarial machine learning at
  scale,'' \emph{arXiv preprint arXiv:1611.01236}, 2016.

\bibitem{deng2009imagenet}
J.~Deng, W.~Dong, R.~Socher, L.-J. Li, K.~Li, and L.~Fei-Fei, ``Imagenet: A
  large-scale hierarchical image database,'' in \emph{Computer Vision and
  Pattern Recognition, 2009. CVPR 2009. IEEE Conference on}.\hskip 1em plus
  0.5em minus 0.4em\relax IEEE, 2009, pp. 248--255.

\bibitem{szegedy2016rethinking}
C.~Szegedy, V.~Vanhoucke, S.~Ioffe, J.~Shlens, and Z.~Wojna, ``Rethinking the
  inception architecture for computer vision,'' in \emph{Proceedings of the
  IEEE Conference on Computer Vision and Pattern Recognition}, 2016, pp.
  2818--2826.

\bibitem{chen2017zoo}
P.-Y. Chen, H.~Zhang, Y.~Sharma, J.~Yi, and C.-J. Hsieh, ``Zoo: Zeroth order
  optimization based black-box attacks to deep neural networks without training
  substitute models,'' \emph{arXiv preprint arXiv:1708.03999}, 2017.

\bibitem{lu2017safetynet}
J.~Lu, T.~Issaranon, and D.~Forsyth, ``Safetynet: Detecting and rejecting
  adversarial examples robustly,'' \emph{arXiv preprint arXiv:1704.00103},
  2017.

\bibitem{metzen2017detecting}
J.~H. Metzen, T.~Genewein, V.~Fischer, and B.~Bischoff, ``On detecting
  adversarial perturbations,'' \emph{arXiv preprint arXiv:1702.04267}, 2017.

\bibitem{carlini2017adversarial}
N.~Carlini and D.~Wagner, ``Adversarial examples are not easily detected:
  Bypassing ten detection methods,'' \emph{arXiv preprint arXiv:1705.07263},
  2017.

\bibitem{meng2017magnet}
D.~Meng and H.~Chen, ``Magnet: a two-pronged defense against adversarial
  examples,'' \emph{arXiv preprint arXiv:1705.09064}, 2017.

\bibitem{bhagoji2017dimensionality}
A.~N. Bhagoji, D.~Cullina, and P.~Mittal, ``Dimensionality reduction as a
  defense against evasion attacks on machine learning classifiers,''
  \emph{arXiv preprint arXiv:1704.02654}, 2017.

\bibitem{lecun1998mnist}
Y.~LeCun, ``The mnist database of handwritten digits,'' \emph{http://yann.
  lecun. com/exdb/mnist/}, 1998.

\bibitem{krizhevsky2009learning}
A.~Krizhevsky and G.~Hinton, ``Learning multiple layers of features from tiny
  images,'' 2009.

\end{thebibliography}

\end{document}